\documentclass[final,5p,times,twocolumn]{elsarticle}

\usepackage{graphicx}
\usepackage{hyperref}
\usepackage[noadjust]{cite}
\usepackage{epsfig}
\usepackage{epstopdf}
\usepackage{bm}
\usepackage[usenames,dvipsnames,svgnames,table]{xcolor}
\usepackage{amsmath}
\usepackage{amssymb}
  
\usepackage{amsthm}
\usepackage{subfig}
\usepackage{textcomp}
\usepackage{times}
\usepackage[T1]{fontenc}
\usepackage{mathtools}
\usepackage{mathrsfs}
\usepackage{booktabs}
\usepackage{multirow}
\usepackage[
singlelinecheck=false,font=small,labelfont=bf
]{caption}
\usepackage{url} 

\usepackage{lineno}

\newcommand{\beq}{\begin{equation}}
\newcommand{\eeq}{\end{equation}}

\makeatletter
\def\ps@pprintTitle{%
 \let\@oddhead\@empty
 \let\@evenhead\@empty
 \def\@oddfoot{\footnotesize\textit{Preprint - Article accepted to Robotics and Autonomous Systems}\hfill\textit{October 1, 2019 }$\blacksquare$}%
 \let\@evenfoot\@oddfoot}
\makeatother

\journal{Robotics and Autonomous Systems}

\begin{document}

\begin{frontmatter}
\title{Multimodal representation models for prediction and control from partial information}
\author{Martina Zambelli}
\ead{m.zambelli13@alumni.imperial.ac.uk}

\author[]{Antoine Cully}
\ead{a.cully@imperial.ac.uk}

\author{Yiannis Demiris}
\ead{y.demiris@imperial.ac.uk}

\address{Personal Robotics Lab, Dept. of Electrical and Electronic Engineering, Imperial College London, UK}
\fntext[label3]{This research was done when M. Zambelli and A. Cully were in the Personal Robotics Lab, Imperial College London. Martina Zambelli's present affiliation is DeepMind London UK. Antoine Cully's present affiliation is the Department of Computing, Imperial College London, UK. }

\begin{abstract}
Similar to humans, robots benefit from interacting with their environment through a number of different sensor modalities, such as vision, touch, sound. However, learning from different sensor modalities is difficult, because the learning model must be able to handle diverse types of signals, and learn a coherent representation even when parts of the sensor inputs are missing. In this paper, a multimodal variational autoencoder is proposed to enable an iCub humanoid robot to learn representations of its sensorimotor capabilities from different sensor modalities. The proposed model is able to
(1) reconstruct missing sensory modalities, (2) predict the sensorimotor state of self and the visual trajectories of other agents actions, and (3) control the agent to imitate an observed visual trajectory. Also, the proposed multimodal variational autoencoder can capture the kinematic redundancy of the robot motion through the learned probability distribution. Training multimodal models is not trivial due to the combinatorial complexity given by the possibility of missing modalities.
We propose a strategy to train multimodal models, which successfully achieves improved performance of different reconstruction models. Finally, extensive experiments have been carried out using an iCub humanoid robot, showing high performance in multiple reconstruction, prediction and imitation tasks.
\end{abstract}


\end{frontmatter}


\section{Introduction}
\label{sec.introduction}

Several studies have revealed that the ability of humans to make predictions is not only essential for motor control, but it is also fundamental for high level cognitive functions including action recognition, understanding, imitation, mental replay, and social cognition \citep{Wolpert2001}.
Improving the ability of robots to make predictions is a promising direction to enhance their skills, not only on motor control and prediction of their own body, but also on fostering the understanding of others' actions.
Well-established learning systems for motor prediction and control~\citep{Wolpert1998,Kawato1999,demiris2006hierarchical}
are built on internal models, namely \textit{forward} and \textit{inverse} models. The former provides a prediction of the state of the agent given the current state and an action, while the latter provides a mapping in the opposite direction: given a target state and the current state, it retrieves the action to bring the system from the current state to the target.
Assuming existing similarities between agents, the internal model used to predict one's own actions can be instrumental to predict the (\textit{visual}) consequences of someone else's actions \citep{demiris2006hierarchical, demiris2014mirror}.
The assumption of the existence of similarities between agents poses a challenge in robotics, known as the correspondence problem \citep{hafner2005interpersonal,alissandrakis2002imitation,nehaniv1998mapping}. This paper does not address this problem. Instead, we assume that the robot has access to  visual information from an egocentric point of view. A solution to address general scenarios where the spatial perspective that the robot acquires of its own and of others' actions is different has been proposed for example in \citep{JohnsonDemiris05, Fischer2016}. However, in this work it is assumed that agents share the same perspective (the same assumption is generally made in similar applications \citep{baraglia2015motor,copete2016motor}).  

\begin{figure*}
\centering
\includegraphics[width=\textwidth]{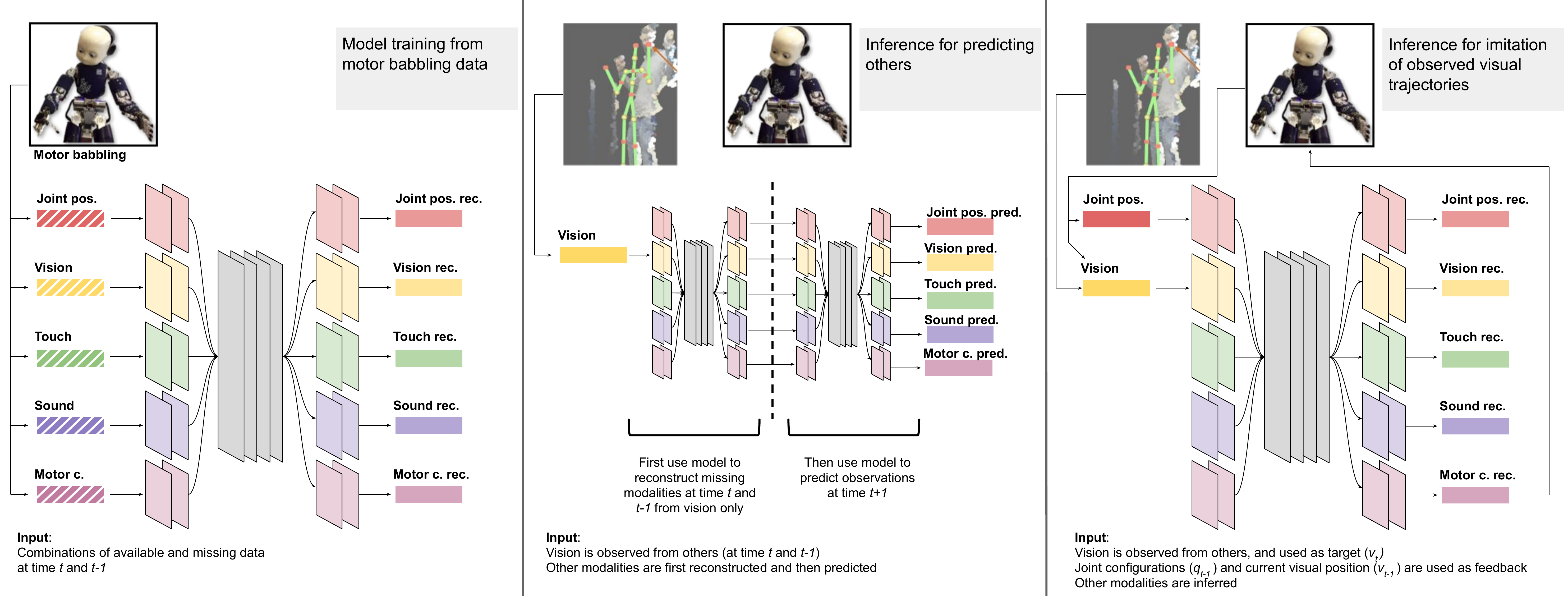}
\caption{Overview of the learning architecture.
The self-learned  model can be used to reconstruct missing data, make predictions, control the robot's motion.
When observing others, only the visual information is available.
The model learned can reconstruct the multimodal state of the robot, including the proprioceptive, visual, tactile, sound and the motor commands data, from partial information (left).
The model can also be used to make predictions in the futures, by feeding reconstructed data back to the model (center).
Finally, the model can generate motor commands that can be issue directly to the robot's joints to imitate others' visual trajectories (right).
}
\label{fig.architecture}
\end{figure*}

While several studies have focused on predicting outcomes of actions of the agent (\textit{e.g.} learning a forward model) or actions of others 
(\textit{e.g.} human trajectories from images or videos)
\citep{kamel2018deep,kamel2019efficient,kamel2019investigation}, 
in this paper the goal is to learn a model of the self that can be applied to  predict and imitate the visual perception of another agent from an egocentric point of view.  
The proposed architecture is based on a self-learned model, which is built, trained and updated only using the experience accumulated by the agent.
The advantage of self-learned models is that they can be used without specific prior knowledge about the robot, for example its morphology or predefined forward and inverse models. This information might be unavailable in some cases, such as in soft robotics or after a mechanical damage. Self-learned models can enable robots to learn on their own how to behave in those circumstances \citep{cully2015robots, kriegman2019automated}.
However, one of the major obstacles in using self-learned internal models to predict motion of others is the intrinsic difference between the available data. While the model is learned and exploited by the agent using a whole range of available sensory modalities, only the visual information is available when observing someone else's motion.
In this paper, we overcome this challenge by implementing  a model
which is able to retrieve the missing sensory information and motor commands needed for  mimicking and predicting the visual trajectories of another agent's action. 
As a result, the main contribution of this paper is a learning architecture that uses 
a multimodal variational autoencoder in a versatile manner to
(1) reconstruct missing sensory modalities, (2) predict the sensorimotor state of self and the visual perception of another agent from an egocentric point of view, and (3) imitate the observed agent's visual trajectory. 
This architecture represents a unified representation of the traditional forward and inverse model leveraging their synergy to implement functions that are fundamental for autonomous systems. An overview of the proposed learning architecture is shown in Fig.~\ref{fig.architecture}. 

Variational autoencoders \citep{kingma2013auto, rezende2014stochastic} have recently emerged as one of the most popular approaches for unsupervised learning of complex distributions of data.
One of their key characteristics is that they can model the probability distribution of the reconstructed data and its distribution in the latent space. 
In this paper, we extend a traditional variational autoencoder model to reconstruct the probability distribution of non-observed modalities (\textit{e.g.} joint positions and velocities) given observed modalities (\textit{e.g.} visual position of the end-effector). Using probability distributions is particularly important in the case of robotics applications, as it allows the system to take into account the redundancy of the system. 
Typically, several joint positions lead to the same end-effector position, and such relationships can be captured by the learned conditional probability distribution. 
An important aspect of this work is also the training strategy used to learn this model. Specifically, we propose to train the model to reconstruct the input even when only part of it is available, by adopting a denoising approach. Our experiments, presented in Section \ref{sec.experiments}, show that this method allows for the improved performance on the task at hand of various alternative models too. 

The paper is organized as follows: the multimodal variational autoencoder implementation is introduced in Section \ref{sec.methodology}.
Experiments have been performed by using a humanoid iCub robot 
and results are reported and discussed in Sections \ref{sec.experiments} and \ref{sec.discussion}, respectively.


\section{Related work}
\label{sec.relatedwork}

\paragraph*{Learning internal models in robotics}
Learning algorithms have proven to be an effective means of building internal models for robots. Learning strategies  achieve flexibility and adaptability in building  robots' kinematic and dynamic models, by incorporating uncertainties and nonlinearities, as well as dynamical changes due to wear, and in limiting the influence of specific engineered settings.
Many approaches to learn controllers for robots have been proposed, including for example reinforcement learning \citep{sutton1998reinforcement,abbeel2007application} and learning by demonstration \citep{argall2009survey,billard2008robot}.
Various implementations have been proposed, such as Gaussian processes \citep{deisenroth2011pilco,williams2009multi}, neural networks \citep{miller1995neural,kawato1988hierarchical} 
and more recently deep neural networks \citep{Hinton2006,levine2016end}.
The majority of these studies have focused on learning controllers, 
where the goal is to learn a policy or an inverse model in order to generate motor commands given a target input. 
Typically, learning forward models has been less investigated in traditional robotics because they can be directly defined based on the kinematic structure of the robot. However, 
learning such models is fundamental to implement a prediction model for robots to be able to make predictions not only on their own actions but also of others' actions.

Forward and inverse models learning is a general approach to allow robots to learn new skills. Forward models generate state predictions from current state and action, while inverse models generate actions from states. These two capabilities enable robots to perform predictions, ``mental simulation'', planning, and control \citep{Wolpert1998,Kawato1999,demiris2006hierarchical}. 
In developmental robotics, such models are acquired by designing learning mechanisms to let a robot build its own perceptive and behavioral repertoire. The focus is to investigate the acquisition of motor skills from sensorimotor interaction with the environment \citep{Lungarella2003}.
As a result, the developmental approach aims to endow robots with all the learning capabilities that may be necessary to build rich and flexible sensorimotor representations
\citep{Sigaud2016}. 
Several studies have addressed the problem of learning internal models from sensorimotor data through exploration strategies, including for example learning of visuomotor models \citep{droniou2012autonomous,vicente2016online}, learning of dynamics models \citep{calandra2015learning}, and learning from multiple sensory signals and possible partial information \citep{fitzpatrick2006reinforcing,vicente2016online,ruesch2008multimodal}.
Internal models (forward and inverse models) are usually learned separately \citep{Wolpert1998,Kawato1999,demiris2006hierarchical}: 
the forward model is used to make predictions, and the inverse model is used for control.
The method proposed in our paper instead achieves these two capabilities in conjunction. This can be a valuable asset, for example in terms of number of parameters used (one network instead of multiple ones). Our proposed approach also provides a compact yet powerful model that can achieve satisfactory performance on both prediction and control tasks.
One powerful way to learn internal models is imitation, considered a fundamental part of learning in humans and used as a mechanism of learning for robots \citep{Demiris2005crib}. 
The ability to predict someone else's movements inherently incorporates the necessity of understanding others' motion, being able to simulate it by developing learning as well as imitation skills.
A vast literature exists in the robotics domain addressing imitation, in particular the paradigm of learning by imitation 
\citep{schaal2003computational,calinon2010probabilistic,lopes2005visual}, 
and the related correspondence problem \citep{hafner2005interpersonal,alissandrakis2002imitation,nehaniv1998mapping} 
arising from the structural (kinematic/dynamic) differences between a demonstrator and a learner agent. Imitation can happen at different levels, such as at the action level, or at the effect level \citep{nehaniv2001like}.
Recently, advances on motion analysis and estimation have been proposed \citep{kamel2018deep,kamel2019efficient,kamel2019investigation}, and these techniques have also been applied to humanoid robot motion learning through sensorimotor representation and physical interactions \citep{shimizu2014robust}.
In this paper, we use a  trajectory level imitation, as an instrumental example of application of our proposed multimodal learning approach. Also, although the correspondence problem has an important role in the context of learning by imitation, we refer the reader to the relevant literature to solve this problem, and we focus the paper on the multimodal learning approach instead.

\paragraph*{Multimodal learning}
In the fields of sensor fusion and pattern recognition, several works have addressed the problem of learning representations from multiple sources, \textit{e.g.} text and audio or text and images~\citep{ramisa2017breakingnews,poria2016fusing}. 
In \citep{Ngiam2011}, a multimodal deep learning approach was proposed, able to cope with data of different types, such as visual and audio data, with cross-modal learning and reconstruction. 
Some work on multimodal learning in robotics was proposed in \citep{zambelli2016,zambelli2016tcds}.
Recent literature has started to address the challenging problem of learning from multiple data sources, using variational inference models (\textit{e.g.} variational autoencoders). Among others, two recent works have shown great potential: the joint multimodal VAE \citep{suzuki2016joint}, and the product-of-experts-based multimodal VAE \citep{wu2018multimodal}. The former learns a joint distribution between two modalities, but trains a new inference network for each multimodal subset, which is generally impractical and arguably intractable. The latter uses a product-of-experts inference network and a sub-sampled training paradigm to solve the multimodal inference problem. Although these methods have been shown to achieve good results in domains such as image processing and text-to-vision tasks, they do not address the problem of multimodal learning from different sensors on a real robot. Such domain is fundamentally different since the data collected by the robot while acting are generally noisy time series of unscaled and heterogeneous data.
The main contributions of our work compared to \citep{suzuki2016joint,wu2018multimodal} are the application domain and the ability of our method to generate actions. Our work is the first, to the best of our knowledge, to use a multimodal formulation of variational autoencoders on a real robotic domain. While in \citep{suzuki2016joint,wu2018multimodal} the addressed domains are purely self-supervised learning applications, not involving actions or control tasks, in this work we successfully use a multimodal VAE model to go beyond self-supervision and achieve imitation, prediction and control tasks.
In \citep{Droniou2015}, an architecture based on deep networks was proposed to make a humanoid robot iCub  learn a task from multiple perceptual modalities (namely proprioception, vision, audio). 
While the method proposed in that paper learns the cross-modal relationships between sensory modalities, it is not able to deal explicitly with missing information. On the contrary, the architecture that we propose here can successfully retrieve missing modalities and use them to both predict and control motion.
Finally, \citep{baraglia2015motor,copete2016motor} have applied deep autoencoders to make a robot predict others' actions through predictive learning, showing how a robot can use a self-acquired model to make predictions of others' goals. In those works, the sequences of signals used for learning are given through kinesthetic teaching. On the contrary, in this paper we use a fully autonomous exploration for the robot to acquire its own sensorimotor data. Furthermore, the variational autoencoder that we propose in this paper is a more general and versatile model for robots to not only predict self and others' motion, but also to perform imitation tasks. It also presents one major advantage compared to the model proposed in \citep{copete2016motor}, namely the ability to capture the redundancy of the robotic system.

\section{Methodology}
\label{sec.methodology}

\subsection{Multimodal variational autoencoder}

A variational autoencoder (VAE) \citep{kingma2013auto, rezende2014stochastic} is a latent variable generative model. It consists of an encoder that maps the input data $x$ into a latent representation $z=\text{encoder}(x)$, and of a decoder that reconstructs the input from the latent code, that is $\hat{x}=\text{decoder}(z)$. Encoder and decoder are neural networks, parameterized by $\theta$ and $\phi$, respectively. The lower-dimensional latent space where $z$ lives is stochastic: the encoder, denoted as $q_\phi (z | x)$ outputs a probability density, generally (as also in our case) a Gaussian distribution. The latent representation $z$ can then be sampled from this distribution. The decoder is denoted as $p_\phi(x| z)$: it gets as input the latent representation $z$ of the input, and outputs parameter of a distribution representing the reconstructed input.
The variational autoencoder model can also be written as  
$p_\theta(x, z) =
p(z)p_\theta(x|z)$ where $p(z)$ is a prior, usually  Gaussian, and $p_\theta(x| z)$ is the decoder.

The information bottleneck given by mapping of the input into a lower-dimensional latent space yields to a loss of information. The reconstruction log-likelihood $\log p_\phi (x| z)$ is a measure of how effectively the decoder has learned to reconstruct an input $x$ given its latent representation $z$.
The training goal is then to maximize the marginal log-likelihood of the data. Because this is intractable~\citep{rezende2014stochastic}, the evidence lower bound (ELBO) is instead optimized, by leveraging the inference network (decoder), $q_\phi(z|x)$, which serves as a tractable distribution. The ELBO is defined as:
\beq
ELBO(x) \triangleq \mathbb{E}_{q_\phi(z|x)}
\left[\lambda log p_\theta(x|z)\right] - \beta KL\left[q_\phi(z|x), p(z)\right]
\eeq
where $KL[p, q]$ is the Kullback-Leibler divergence between distributions $p$ and $q$, while $\lambda$ \citep{wu2018multimodal} and $\beta$ \citep{higgins2016beta} are parameters balancing the terms in the ELBO. The ELBO is then optimized via stochastic gradient descent, using the reparameterization trick to estimate the gradient \citep{kingma2013auto, rezende2014stochastic}.
In practice, since the main focus of this study is the reconstruction capability of the model, we chose $\beta=0$, and only consider the reconstruction loss  to train our architecture, noticing improvements in the reconstruction performance obtained.

\begin{figure}[ht]
\centering
\includegraphics[width=.9\columnwidth]{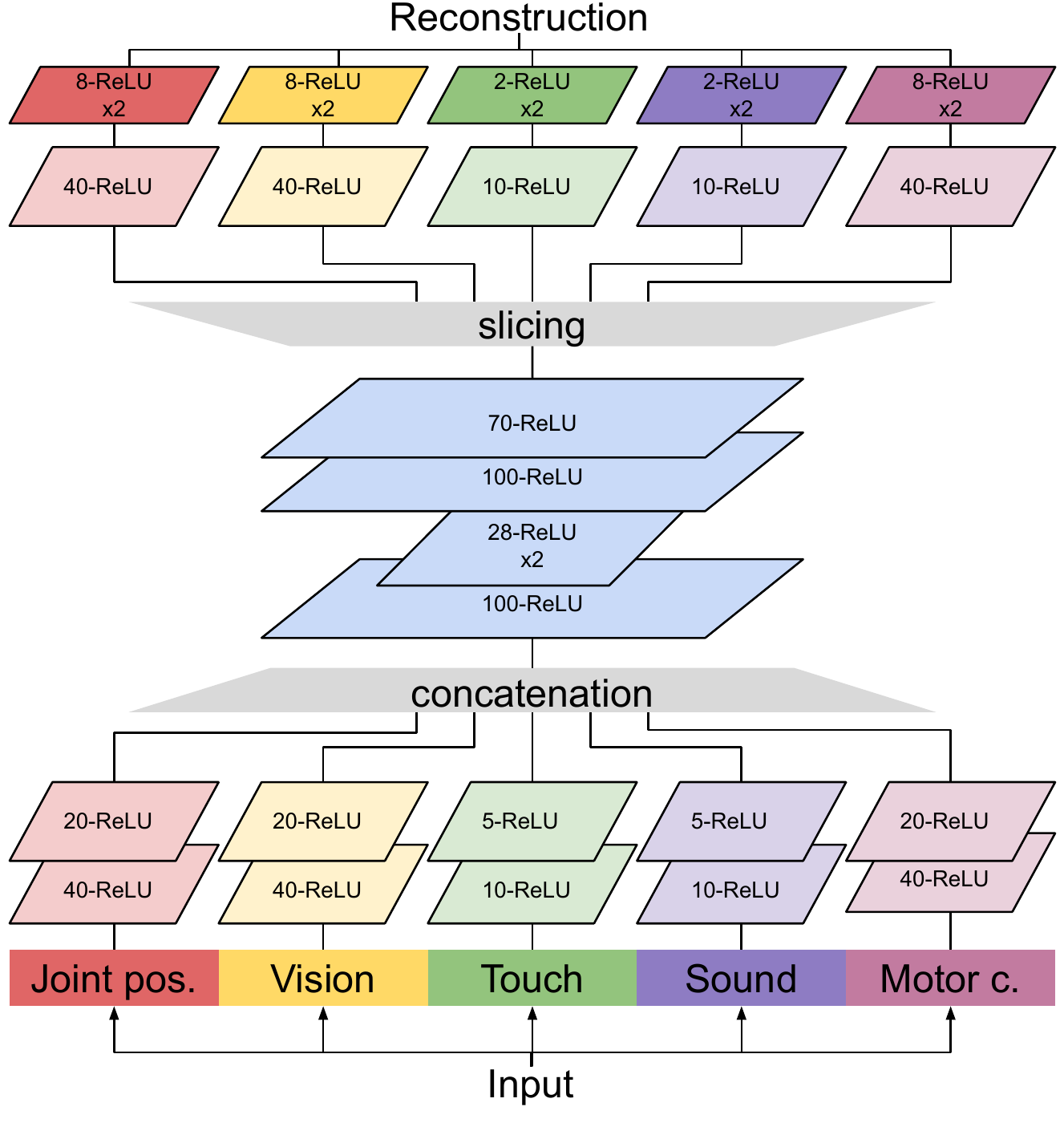} 
\caption{Multimodal Variational Autoencoder used in this work. The input layer is composed by multimodal sensorimotor data. Each modality is encoded and decoded by separate autoencoders (shown with different colors). A shared layer (in light blue, in the center) allows to learn a shared representation among different dimensions. This architecture is trained with complete as well as partial data (see Table~\ref{tab.training_struc}). Each uni-modal autoencoder can be trained separately, allowing for single modality learning. The cross-modality representations are also learned through the shared layer. The output of the network consists of the mean and variance of the reconstruction of each different data part. Details about the parameters of the network included in this figure are further explained in Appendix. N-ReLU represents a fully connected layer with N neurons and using the ReLU activation function. N-ReLU x2 indicates that 2 N-ReLU layers are created in parallel, one to encode the mean and the other to encode the variance of the output distribution. }
\label{fig.mdvae}
\end{figure}

In this paper, we extend standard variational autoencoders to multimodal sensorimotor data.
Our multimodal VAE is formed of multiple encoders and decoders, one for each sensory modality. Each encoder and decoder is an independent neural network, not sharing weights with other modalities' networks. The latent representation is however shared: each encoder maps its input (one sensory modality) into the shared code $z$, as depicted in Fig.~\ref{fig.mdvae}. Each decoder then reconstruct its particular output (one sensory modality) from the shared code.
The main difference that characterizes the \textit{multimodal} learning approach compared to a standard VAE is that the sub-networks can be used to process each modality, and shared layers can be used to learn cross-modal relations (see Fig. \ref{fig.mdvae}).

The $\lambda$ parameters are here used to balance losses from different sensor modalities.
In order to put more emphasis on modalities  described by fewer dimensions (\textit{e.g.} the tactile and sound modalities), 
we compute independent loss values for each modality ($m$) and weight them according to their dimensionality ($D_m$), that is $\lambda_m = 1/D_m$. 
Then the sum of the independent reconstruction loss terms is optimized.
The scaling factor given by the dimensionality of each modality allows us to balance the importance of each modality when combining them in the optimization step. That is, when optimizing the reconstruction loss, the $\lambda$ weights allow to take into account that each modality and each corresponding unimodal sub-network have different dimensions.
This approach helps learning even the most difficult parts of the state space, such as discrete or binary dimensions of the sensory space (see tactile example in Figure~\ref{fig.data_self}).

This type of variational model presents various advantages in a robotic framework.
First, the ability of variational autoencoders to learn the distribution of a dataset in latent space is a powerful feature to generate a shared representation of the different modalities. For instance, the latent representation can be used to learn relationships and dependencies present in the sensorimotor experience of robots. This can be leveraged to generate new artificial perception by sampling from the latent distribution in the latent space. 
Second, this shared latent representation also allows the robot to reconstruct missing modalities. For example, if data from a sensor is unavailable, this model can be used to model the probability distribution of the data that should be observed from this sensor conditioned on the data from other sensors of the robot. 
Finally, their ability to predict probability distributions is fundamental to take into account the redundancy of complex robots, such as the iCub humanoid robot used in this study. With this property, the model can capture the fact that for a given end-effector position, several joint configurations are possible.   

Details of the network implemented and used in this work are reported in~\ref{app.architecture}.

\subsection{Training the Multimodal Variational Autoencoder}
\label{sec.meth_train}

An important contribution of this work is the training strategy used to learn the proposed model. We propose to train the model to reconstruct the input even when only part of it is available, by adopting a de-noising approach. While in the following paragraphs the proposed training approach is presented relative to the multimodal variational autoencoder introduced earlier, this strategy is generic, and can be applied to other architectures, such as the reconstruction model proposed in \citep{Droniou2015} as demonstrated by the experimental results.
In the Experiment section, we show that the proposed training strategy allows to improve performance on the task at hand of various alternative models. 

The training dataset contains multimodal sensorimotor data collected during a self-exploration phase. Data are captured from different sensors of the robot, such as the position of the hand in the robot's visual space, tactile and sound data, and  proprioception (joint positions) from the motor encoders. In particular, the position of the hand in the visual space is extracted by considering the center point of a tracking window around the moving hand. All data are then normalized to take values in the range $[-1,1]$.  More details regarding the data acquisition and the database are presented in Section \ref{sec.experiments_setup}.

\begin{figure}
\centering
\includegraphics[width=\columnwidth]{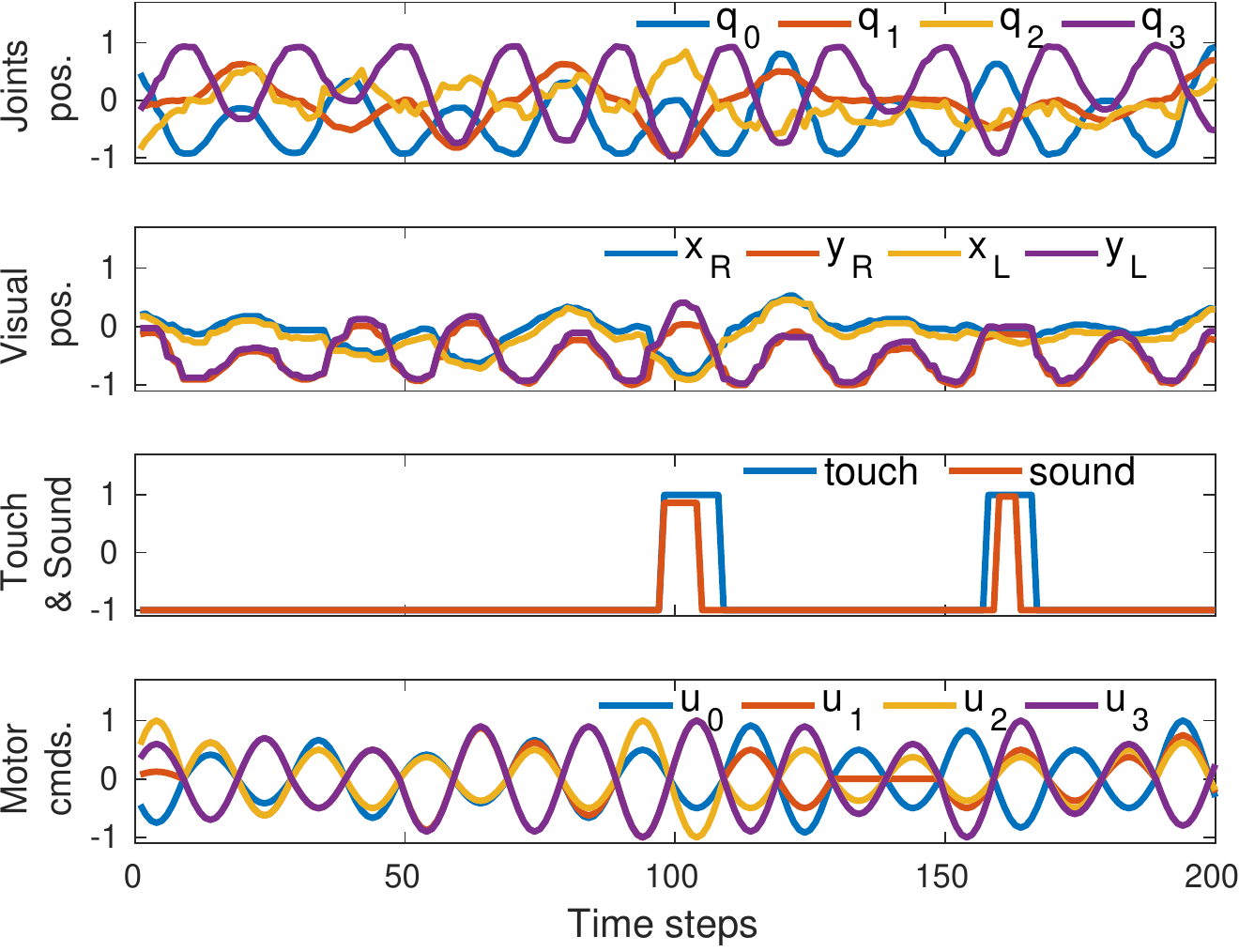}
\caption{Data from self-exploration.
Joint positions are recorded from the motor encoders, visual positions (4D) are acquired from the RGB cameras of the robot's eyes, random sinusoidal velocity commands are issued to the arm joints to perform motor babbling.
\label{fig.data_self}}
\end{figure}

Time series data from the self-exploration dataset recorded are shown in Fig.~\ref{fig.data_self}. Denote by ${\mathbf{u}_t}$ the vector of velocity commands issued at time $t$, ${\mathbf{q}_t}$ the vector of joint positions (proprioception), ${\mathbf{v}_t}$ the vector of the visual position, $\mathbf{p}_t$ the tactile signal and $\mathbf{s}_t$ the sound signal at time $t$. Note that other modalities can also be included. The input of the architecture is a multi-dimensional vector ${\mathbf{y}_t=[\mathbf{q}_t,\mathbf{q}_{t-1},\mathbf{v}_t,\mathbf{v}_{t-1},\mathbf{p}_t,\mathbf{p}_{t-1},\mathbf{s}_t,\mathbf{s}_{t-1},\mathbf{u}_t,\mathbf{u}_{t-1}]}$, which contains both data from time $t$ and $t-1$ to capture the temporal relationship between the different modalities. 

The network is trained on both complete and partial samples of the training dataset collected during the robot self-exploration. To do so, the original dataset is augmented with samples that require the network to reconstruct the missing modalities given only one of them. This is realized by duplicating the dataset, while using a flag value (namely the arbitrary value -2, which is outside the range of any sensorimotor signal after normalization) to denote the non-observable modalities. The training dataset follows the structure in Table~\ref{tab.training_struc} to enable the network to perform predictions and reconstruction in multiple conditions of missing information.
More specifically, the augmented training set is formed by concatenating the original complete set of data collected during motor babbling and normalized to values between -1 and 1, with mutilated versions of itself. The final dataset is then (1) the complete data at time $t$ and $t-1$, concatenated to (2) data including only time $t-1$, concatenated to (3) data including only proprioception at time $t-1$ and vision at time $t$ and $t-1$, concatenated to (4) data including only vision at $t$ and $t-1$.
At each training step, a batch is randomly sampled from the augmented dataset and fed to the multimodal VAE model. The batch may contain only partial data, but the training objective forces the network to try to reconstruct the target complete sensorimotor state (\textit{i.e.} $\mathbf{y}_t$). 
Because the model is trained using the combination of complete and partial data as described above, the latent representation is shaped in such a way that it is robust to missing data; similarly, the sub-networks weights are learned to also be robust to missing inputs.

\begin{table}
\centering
\caption{Training dataset structure: original dataset (1) augmented with samples that only include partial data (2-3-4). Each row correspond to a dataset of 7380 datapoints. Colored cells indicate that the corresponding modality is present in the dataset. For the cases (2-3-4), missing modality data (cells in gray) is replaced with values $-2$. The datasets (1), (2), (3), and (4) are concatenated. The proposed model is trained on the augmented dataset, that is the concatenation of the four (1-2-3-4).
\label{tab.training_struc}}
\resizebox{\columnwidth}{!}{%
\begin{tabular}{| l || c | c | c | c | c | c | c | c | c | c |}
\hline 
(1) & \cellcolor{red!35}$\mathbf{q_t}$ & \cellcolor{red!35}$\mathbf{q_{t-1}}$ & \cellcolor{yellow!35}$\mathbf{v_t}$ & \cellcolor{yellow!35}$\mathbf{v_{t-1}}$ & \cellcolor{green!35}$\mathbf{p_t}$ & \cellcolor{green!35}$\mathbf{p_{t-1}}$ & \cellcolor{blue!35}$\mathbf{s_t}$ & \cellcolor{blue!35}$\mathbf{s_{t-1}}$ & \cellcolor{purple!35}$\mathbf{u_t}$ & \cellcolor{purple!35}$\mathbf{u_{t-1}}$ \\ \hline
(2)&\cellcolor{gray!15}- & \cellcolor{red!35}$\mathbf{q}_{t-1}$ & \cellcolor{gray!15}- & \cellcolor{yellow!35}$\mathbf{v}_{t-1}$& \cellcolor{gray!15}-  & \cellcolor{green!35}$\mathbf{p}_{t-1}$& \cellcolor{gray!15}-  & \cellcolor{blue!35}$\mathbf{s}_{t-1}$& \cellcolor{gray!15}-  &\cellcolor{purple!35} $\mathbf{u}_{t-1}$ \\\hline
(3)&\cellcolor{gray!15}- &\cellcolor{red!35} $\mathbf{q}_{t-1}$ & \cellcolor{yellow!35}$\mathbf{v}_t$ & \cellcolor{yellow!35}$\mathbf{v}_{t-1}$&  \cellcolor{gray!15}-  &  \cellcolor{gray!15}- &  \cellcolor{gray!15}- &  \cellcolor{gray!15}- &  \cellcolor{gray!15}- &  \cellcolor{gray!15}- \\\hline 
(4)&\cellcolor{gray!15}- & \cellcolor{gray!15}- & \cellcolor{yellow!35}$\mathbf{v}_t$ & \cellcolor{yellow!35}$\mathbf{v}_{t-1}$& \cellcolor{gray!15} -  & \cellcolor{gray!15} - & \cellcolor{gray!15} - & \cellcolor{gray!15} - & \cellcolor{gray!15} - & \cellcolor{gray!15} - \\  
\hline 
\end{tabular}
}
\end{table}

 \subsection{One model for multiple tasks} 
One of the major assets of our proposed model is its versatility, that is the possibility of using the same learned model to achieve different goals. In this section, we present how the learned  multimodal variational autoencoder can be deployed to achieve three different objectives:
\begin{enumerate}
\item reconstructing missing data;
\item predicting the robot's own sensorimotor data and visual trajectories from other data sources (\textit{e.g.} other agents, other datasets);
\item controlling the robot in an online control loop.
\end{enumerate}
In these three cases, the training, structure, and parameters of the neural network remain the same: the learned model and network used for learning \textit{do not change} even when different sets of input are available. We argue that this is a key aspect of our method: one single model can be trained and learned to capture a comprehensive internal model from multimodal data, and to cope even when part of this data is not available. 
Details for each of the aforementioned functions that the model can achieve are given in the remaining part of the section.

\subsubsection{Reconstructing missing data}
Similar to denoising autoencoders, the proposed multimodal VAE is trained to reconstruct missing data.
Missing modalities are set to $-2$ (as explained in Section~\ref{sec.meth_train}), while the network outputs the probability distribution of the reconstructed inputs. 
This is fundamental to address the problem at the origin of this work, that is the ability to predict  the visual trajectory of others taken from  egocentric visual information 
by relying on internal models of the self. In such an application, an agent learns internal representations of its sensorimotor space, in particular relating motor actions with multimodal sensory effects \citep{demiris2006hierarchical, demiris2014mirror, Pickering2014}. However, when observing someone else performing an action, only the visual information is available. The agent, which relies on full information from all its sensors, must then be able to retrieve the missing information and interpret the observed motion in relation with its own internal representations. The architecture proposed in this paper allows robots to achieve this by reconstructing the missing sensorimotor information; for example reconstructing joint configuration, touch, sound and motor information from  observations of the visual input, or time step $t$ from observations at time $t-1$.

\subsubsection{Predicting the robot and others'  visual trajectories  }

While data from all  sensory modalities is available to the agent when learning the models, only the visual input, from an ego-centric perspective is available when observing others.
This implies that only data referring to the visual input are available in $\mathbf{y}$ (see (4) in Table~\ref{tab.training_struc}: this part of the augmented dataset only contains visual data at time $t$ and $t-1$; training on this part of the dataset allows the network to learn to predict the missing modalities from only visual information).

In this respect, the reconstruction of missing modalities described above plays a key role. The neural network can act as a forward model to predict the next sensorimotor state $\mathbf{y_{t+1}}$ from the current state of the agent  $\mathbf{y_{t}}$ (see line (2) in Table~\ref{tab.training_struc}: this part of the augmented dataset only contains data at time $t-1$; training on this part of the dataset allows the network to learn to predict the next time step when only the previous observation is available). However, when observing someone else, the current state of the agent is not fully available, as only vision information can be observed. 
To perform predictions the network needs to infer future sensorimotor states given the current one first. 
We first feed the model with $\mathbf{y_{t-1}}$, and let the model reconstruct $\mathbf{y_{t}}$; then we feed the obtained reconstructed signal $\mathbf{y_{t}}$ as if it was the $t-1$ observation, and let the network reconstruct the missing part, that is $\mathbf{y_{t+1}}$.
In summary, the network first reconstructs the current sensorimotor perceptions of the observed agent and then uses these reconstructed perceptions to predict the next state of the agent.

\subsubsection{Controlling the robot in an online control loop}
In addition to the abilities of the architecture to reconstruct and predict the visual  trajectories of other agents' motion, the learned model can be used as a controller for the robot. In particular, we show how the model can be placed in a control loop to regulate the sensory state of the robot given a target state. This approach can be used in imitation learning scenarios, for instance, where the robot imitates a target trajectory. In our scenario, the robot observes someone else's visual trajectory from an egocentric point of view and uses the learned model to replicate such trajectory.

The control loop is depicted in Fig.~\ref{fig.architecture} (rightmost diagram). 
Notably, the joint and visual configurations ($\mathbf{q_{t-1}},\mathbf{v_{t-1}}$) of the robot are fed back to the network in order to provide the correct current state at each time. This prevents the network from drifting during the online cycles of the control loop, due to the dependencies between different input modalities. For example, moving to areas of the sensory space that lie far from the training space have increased uncertainty. This condition is made more severe by the multimodal nature of the data, which come independently from diverse sensors. The feedback loop implemented to provide the network with the real current data from the robot helps prevent the accumulation of errors in different state dimensions.

It is also important to emphasize that using the learned network as a controller for the robot is not a trivial application, since the network itself represents a model of the robotic system. The ability of the network to produce motor commands is then key to achieve a controller behavior, but this is not sufficient to implement an effective controller. It is important to provide the network with all the sensory information that can help the model to learn the kinematics and dynamics of the system, in particular the sensory states at two consecutive time steps. This is key for the network to build meaningful representations of the robot kinematics and dynamics, and in turn to generate sensible motor commands.

\section{Experiments}
\label{sec.experiments}

\subsection{Experimental setup}
\label{sec.experiments_setup}
We have demonstrated our proposed approach using a humanoid iCub robot. In our scenario, the robot is interacting with a piano keyboard.
The architecture is trained using data collected from the robot 
through experience, by performing pseudo-random self-exploratory movements (motor babbling). 

Then, the robot uses the learned architecture to (1) reconstruct missing sensory modalities, (2) predict  its own sensorimotor state and predict visual trajectories of another agent from an egocentric point of view, and (3) imitate the observed agent 's trajectories.

During the experiments, the iCub robot moves it's right arm, while keeping its head still to a fixed position. 
Four joints of one of the robot's arms are used during motor babbling. 
 The joints' positions ($q_0,...,q_3$) are acquired from the motor encoders attached to each joint\footnote{The initial joints configuration of the robot's arm is $q_0$=-35 deg,  $q_1$=35 deg, $q_2$=0 deg, $q_3$=50 deg (with $q_0,...,q_3$ corresponding to the shoulder pitch, roll, yaw, and elbow flexion, respectively), the wrist is fixed in the standard neutral position, the index finger extended in the neutral position and the rest of the fingers folded. The joint configuration of the robot's head is the standard neutral one, except for the first two joints of the neck which are turned 12 degrees rightwards and downwards.}.
Visual information encoding the position of the hand in the 2D visual field of the robot is acquired from the robot's eye cameras 
(using a resolution of $320\times 240$ pixels for the image frames), 
with coordinates $x_R,y_R$ and $x_L,y_L$ for the right and left eye, respectively.  This is obtained by tracking the hand of the robot using OpenCV features and computing the mean of the tracked feature points, thus obtaining the two coordinates in the 2D frames. This approach is a coarse representation of the visual information available to the robot. An alternative is to extract visual information directly from pixels using a convolutional neural network (CNN). On the other hand, the coarse approximation obtained with the simple visual tracker was sufficient to develop the experiments presented in the following paragraphs, and we let the implementation of the CNN as future work.   
A binary one-dimensional tactile signal is acquired from the robot's artificial skin, which consists of a network of taxels (``tactile pixels'').  
More specifically, the 60 tactile signals acquired from the robot's hand skin are normalized, averaged and binarized using an empirically fixed threshold. The result is a one-dimensional signal that is equal to 1 when a contact is perceived (\textit{i.e.} when the average of the signals is above the fixed threshold), or 0 otherwise. 
Sound data is acquired from the piano keyboard, in the form of a one-dimensional vector containing the MIDI information related to the key played.  MIDI is a symbolic representation of musical information incorporating both timing and velocity for each note played, which is thus associated to a specific integer number. 
The commands sent to the robot's motors ($u_0,...,u_3$) to perform autonomous self-exploration (motor babbling) are velocity references. 
No prior knowledge is assumed on the robot's kinematic or dynamic structure. The choice of using velocity commands aims to keep this prior knowledge to a minimum by avoiding to rely on the inverse kinematic of the robot.  
However, our method can accommodate other implementation choices, such as position or torque control.
Self-exploration is realized by performing motor babbling on one of the robot's arm.
Random sinusoidal motor commands are sent to the motors as velocity commands defined for each joint $j$ as 
$
u_j(t) = \alpha_j \sin (2\pi\omega t ),
$
where the amplitudes $\alpha_j$ are sampled for each joint at each cycle from a uniform distribution $\mathcal{U}(-\bar{u},\bar{u})$, and the frequency $\omega$ is fixed so that each cycle starts and terminates at zero (\textit{i.e.} null velocity).
Normalization is finally applied to all data to obtain signals in the range $[-1,1]$.
 The dataset collected from motor babbling contains 7380 data points, corresponding to approximately 30 minutes of exploration. This dataset is then augmented in order to train the network, as explained in the previous section. 

The input fed to the network is a 28-dimensional vector, including two four-dimensional joint position vectors ($\mathbf{q_t,q_{t-1}}$), two four-dimensional visual position vectors ($\mathbf{v_t,v_{t-1}}$), two one-dimensional tactile vectors ($\mathbf{p_t,p_{t-1}}$), two one-dimensional sound vectors ($\mathbf{s_t,s_{t-1}}$), and two four-dimensional motor commands vectors ($\mathbf{u_t,u_{t-1}}$).

We performed extensive evaluation tests of our proposed method. Three different datasets have been used: test data from the robot self-exploration,  data from a RGB-D camera of a human playing a piano keyboard, and  data from a RGB-D camera of the Imperial-PRL KSC Dataset\footnote{Dataset available at www.imperial.ac.uk/PersonalRobotics} (data used in \citep{chang2017learning} to validate kinematic structure correspondences methods).
To demonstrate our proposed method in practice, we show that the iCub robot is able to leverage its prediction capability to plan its own actions to imitate a human on the piano keyboard. 

More details about the datasets used (including number of datapoints and training specifications) are provided in~\ref{app.datatrain}.

\begin{figure*}[!h]
\centering
\includegraphics[width=0.95\textwidth]{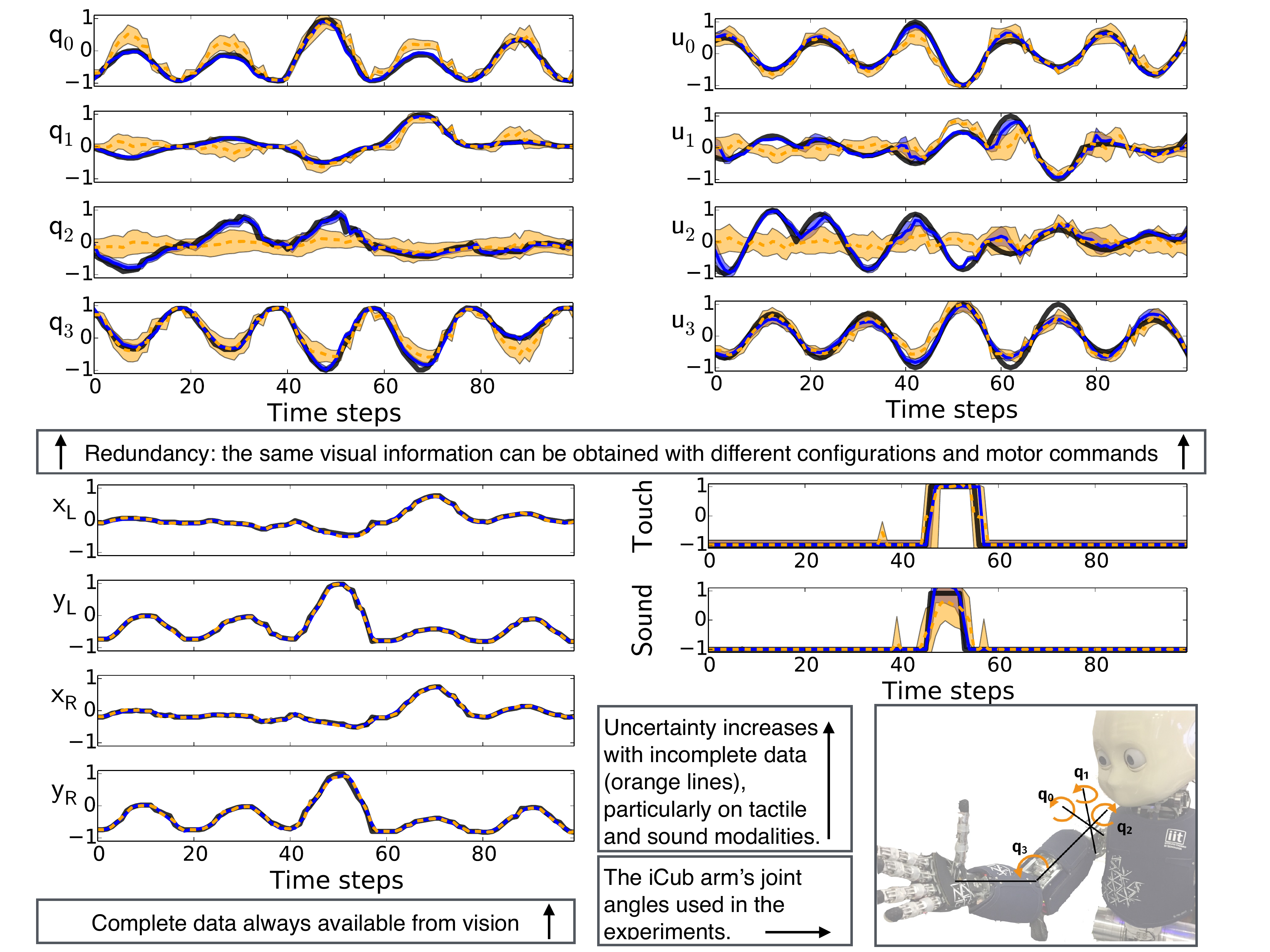}
\caption{Reconstruction results of multimodal data: proprioception ($q_0,...,q_3$), vision ($x_L, y_L, x_R, y_R$), motor commands ($u_0,...,u_3$), touch and sound. Blue lines show the reconstructed data given complete input (case 1 in Table~\ref{tab.training_struc}), and orange lines show the reconstruction results with partial input (case 4 in Table~\ref{tab.training_struc}). Shaded areas represent the variance of the predicted Gaussian distribution of the reconstructed signals.  The multimodal variational autoencoder is able to reconstruct the visual position accurately. Reconstruction results on the joint and motor spaces display the effect of the redundancy of the robot's arm: the same visual position can be reconstructed using diverse configurations, and applying diverse motor commands.
Reconstruction errors occur simultaneously on different degrees of freedom, according to the robot's kinematic structure. The redundancy effect is particularly evident for the second and third joints (${q_1, q_2}$).
A representation of the degrees of freedom of the iCub arm is depicted in the lower-right picture.
}
\label{fig.vae-redundancy}
\end{figure*}

\subsection{Architecture structure}
The network implemented\footnote{Tensorflow \citep{tensorflow2015-whitepaper} has been used for the implementation of the Multimodal Variational Autoencoder.} consists of five unimodal sub-networks, for the proprioceptive (joint positions), visual, tactile, sound and motor modalities, respectively. The encoders, one for each unimodal sub-network, consist of two  fully connected  layers, while the decoders consist of three  fully connected  layers. 
For the proprioception, visual and motor networks, the two encoder layers consist of 40 and 20 units, respectively, and the three decoder layers consist of 40, 8 and 8 units. For the tactile and sound networks, the two encoder layers consist of 10 and 5 units, respectively, and the three decoder layers consist of 10, 2 and 2 units. 
 The ReLU activation function is used throughout the network for each layer. 
The difference in the number of units is to take into account that tactile and sound data are two-dimensional vectors, while the other modalities consist of eight-dimensional vectors.
The outputs of all the unimodal encoders are concatenated to feed into the shared network, which consists of a two-layer encoder with 100 and 28 units, and a two-layer decoder with 100 and 70 units \footnote{
The source code and the dataset used for this experiment can be downloaded at \url{github.com/ImperialCollegeLondon/Zambelli2019_RAS_multimodal_VAE}. }.

\subsection{Sensorimotor data reconstruction}

 In this section, we present experiments that demonstrate the performance of the proposed system to reconstruct sensorimotor data from complete and from partial observations, that is when all inputs are available and when only a subset of modalities is available. The experiments show that the proposed architecture can effectively reconstruct the data in all cases. 

The Multimodal Variational Autoencoder is first trained using datapoints explored during motor babbling.
The dataset collected during babbling is split into a training dataset and a test dataset. 
As described in Section~\ref{sec.methodology}, the network is trained on both complete and partial data of the training set.

In order to evaluate the reconstruction ability of the network, we first assess whether the encoding and decoding of the variational autoencoder manage to retrieve complete input data (when all the modalities are present).
Then we tested the model on the reconstruction of missing modalities, using only the visual information as input.

The experiments conducted showed that the learned network achieves considerable results in terms of reconstruction and beyond that in terms of capturing the complexity of the system.
The network is able to provide an estimate of the input reconstructed even when the majority of the modality dimensions is missing. Importantly, the model is also able to provide a measure of the uncertainty due, for example, to the redundancy of the system.
Results of the reconstruction obtained using the multimodal variational autoencoder are shown in Fig.~\ref{fig.vae-redundancy}.
This figure shows the reconstruction results on the joints, motor, tactile, sound and visual spaces, obtained with both complete and partial input data. 
The data used for this experiment belongs to the dataset collected from the robot self-exploration phase, but have not been used during the training of the model.
It is possible to note that while the reconstruction of the visual signals is very accurate, the reconstruction of the joints' positions and of the motor commands presents a peculiar behavior. In particular, reconstruction errors occur simultaneously for diverse joints. 
A closer analysis of these results shows that these joints are actually related in the kinematic structure of the robot: one joint can compensate or contribute for the movements of the other joint.

The results shown in Fig.~\ref{fig.vae-redundancy} demonstrate how this redundancy is captured by the multimodal variational autoencoder, thus demonstrating the power of this type of network on such difficult tasks. More specifically, the multimodal variational autoencoder is able to learn the general sensorimotor structure underlying the robot's movements rather than single trajectories or single motion sequences. In other words, a robot learns that there can be diverse configurations to achieve a target (for example a visual target). 
For instance, it can be seen that for $q_1$ and $q_2$ the variance of the reconstruction is particularly large. This comes from the fact that several joint configurations can explain the visual information provided to the architecture.
We  note that the true data to be reconstructed remains most of the time within the confidence range of the reconstruction.

The results obtained show another interesting capability of the learned network, namely the ability of learning a forward kinematics only using 2D images from the robot's cameras, while not having direct access to the  depth information of the  3D position of the hand in the robot's operational space. This allows the system to avoid the use of stereo vision algorithms (with the related calibration and matching issues), while having the possibility to rely on the on-board 2D RGB cameras. 

The mean squared errors of the reconstructed sensorimotor signals on test data for each modality have been computed to provide a quantitative account of the network performance.
In Table~\ref{tab.rec}, we report the error scores obtained both when complete and partial data are provided to the network.
Note that the error scores achieved with partial data are comparable to those obtained when feeding complete data to the network, 
 with the only exception of the touch modality, which remains a challenge due to its binary nature. 
This shows that the performance of the network is generally not degraded significantly when the input data consists only of partial data (\textit{i.e.} vision only).
This also shows that the network has successfully learned not only a direct reconstruction of each single modality but also  cross-relations between the modalities and the way to reconstruct one of them provided  only visual data are available.

The values reported in Table~\ref{tab.rec} show the accuracy of the proposed method. The values are reported with percentages (relative to the dataset ranges) to enable direct comparison across the modalities. However, to better appreciate them, consider that the $0.46\%$ and $1.39\%$ mean squared errors in joint space correspond to mean errors of $1.29$ and $2.24$ degrees in joint angles respectively. Similarly, the $0.05\%$ mean squared error in vision space corresponds to an average error of about $1.85$ pixels in the original image frames and mean squared errors of $1.29\%$ and $2.32\%$ in the motor commands represent an average error of $2.16$ and $2.89$ degrees per second.

\begin{table}
\centering
\caption{ Mean squared error percentages for each dimension of the multimodal reconstructed signal on test data.  
\label{tab.rec}}
\resizebox{\columnwidth}{!}{
\begin{tabular}{l | c | c}
& Rec. complete data &Rec. partial data\\
\hline
$\mathbf{q}$ & 0.46\% [0.45; 0.48]\% & 1.39\% [1.37; 1.44]\%\\
$\mathbf{v}$ & 0.05\% [0.04; 0.07]\% & 0.05\% [0.03; 0.06]\%\\
$\mathbf{p}$ & 2.35\% [1.74; 3.66]\% & 9.42\% [9.07; 10.44]\%\\
$\mathbf{s}$ & 3.35\% [0.70; 4.18]\% & 3.95\% [3.35; 4.18]\%\\
$\mathbf{u}$ & 1.29\% [0.67; 1.60]\% & 2.32\% [2.29; 2.37]\%\\
[0.5ex] \hline
\end{tabular}
}
\end{table}

\subsection{Predict own sensorimotor states and visual trajectories of others}
In this section, we present experiments which demonstrate the ability of the proposed architecture to predict the robot's own sensorimotor state and to predict visual trajectories of another agent from an egocentric point of view. These experiments show that the proposed architecture can effectively predict future states by using the multimodal representations learned during training. 
Condition (2) in Table~\ref{tab.training_struc} was critical to achieve this behavior.
The prediction tasks requires the network to infer future sensorimotor states given the current one (see case (2) in Table~\ref{tab.training_struc}). This is realized by feeding the inferred missing time step (\textit{i.e.} the time step $t$) back to the network as the new time step $t-1$, letting the network infer the new time step $t$, which is in fact the prediction at $t+1$.

\begin{figure}[pt]
\centering
\includegraphics[width=\columnwidth]{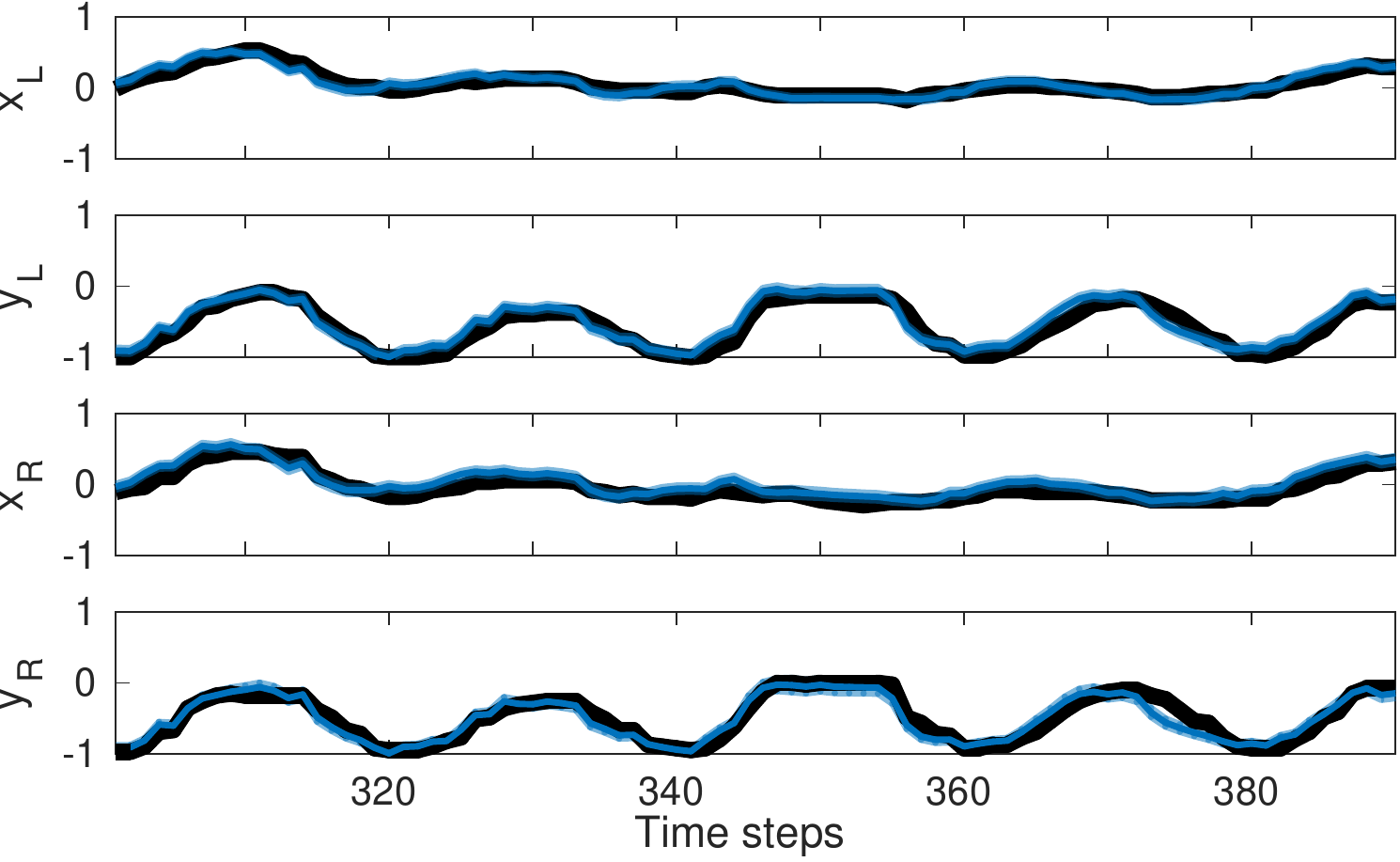}
\caption{
Prediction results using the learned model to predict the visual trajectories (with coordinates $x_L,y_L,x_R,y_R$) of the robot's own motion (a representative part of the trajectories is depicted).
Solid black lines represent the real data (part of the test database), while  blue lines represent the predicted mean and the shaded light blue areas the predicted variance (uncertainty) of the model.
In each plot, on the horizontal axes are the time steps, while on the vertical axes are the magnitude (normalized) of each of the four dimensions of the visual state.
\label{fig.self_pred}}
\end{figure}

First, we have evaluated the proposed architecture using test data from the robot's own data collected from motor babbling.
Results of the predictions of the visual trajectories obtained on data explored during motor babbling are shown in Fig.~\ref{fig.self_pred}.
The mean squared prediction error score obtained on this experiment is  $0.21\%$ (corresponding to less than 4 pixels).  The data on which the experiment is carried out is the test database, that is a part of the data from the robot's self-exploration which was not used for training the model.
These results show that the network is able to effectively make accurate predictions by first reconstructing missing data from visual positions only, and then iterating the process for a second time in order to achieve the next step prediction.

\begin{figure}
\centering
\includegraphics[width=\columnwidth]{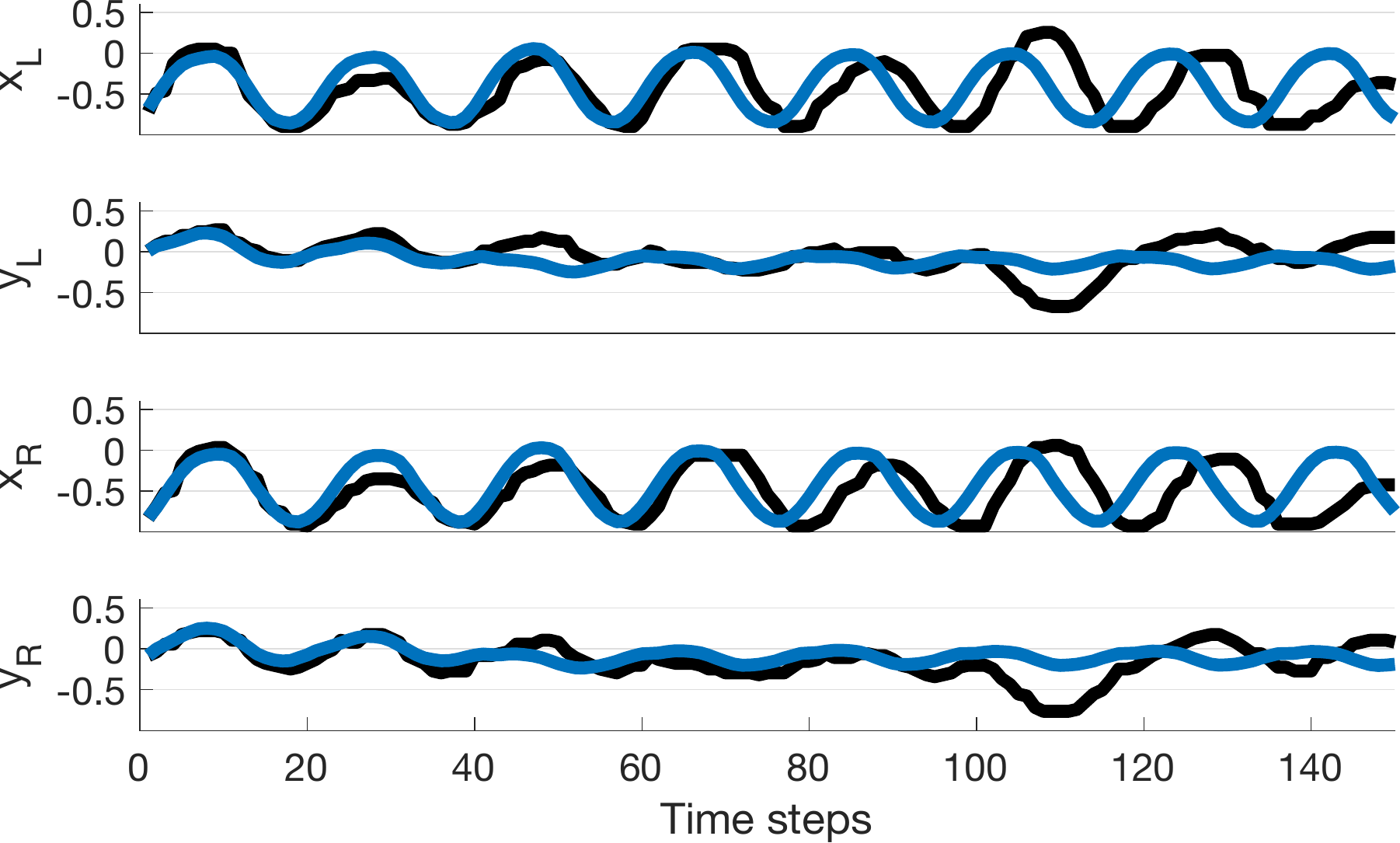}
\caption{
Prediction over multiple time steps using the learned model to predict the visual trajectories (with coordinates $x_L,y_L,x_R,y_R$) of the robot's own motion (a representative part of the trajectories is depicted). 
Solid black lines represent the real data, while  blue lines represent the predicted mean.
\label{fig.multistep}}
\end{figure}

We have also tested the architecture on multi-step ahead predictions. At each time step, the predicted next state is used as the input of the network to predict an additional step ahead. This process can be repeated as long as necessary. The results in Fig.~\ref{fig.multistep} show that the model is capable of predicting the visual trajectory of the on-going swing of the robot (the starting state of the prediction being 2 time steps after the beginning of the swing). The predicted trajectory (in blue) matches accurately the ground truth trajectory (in black) for more than 20 time steps. The prediction accuracy at 50 time steps is $0.42\%$ (less than $5$ pixels).
Then, the model converges to a stable periodic swing pattern which differs from the actual trajectory of the robot. 
Note that obtaining stable long-term predictions with this type of approach is a challenging problem: this approach tends to diverge quickly because of the accumulation of error; also, note that it is expected the model to be unable to predict the movements of the robot after the first swing, as each swing is independent.

\begin{figure}
\centering
\subfloat[Human piano playing.]{\label{fig.martinaPianokinectdata} \includegraphics[height=2.8cm]{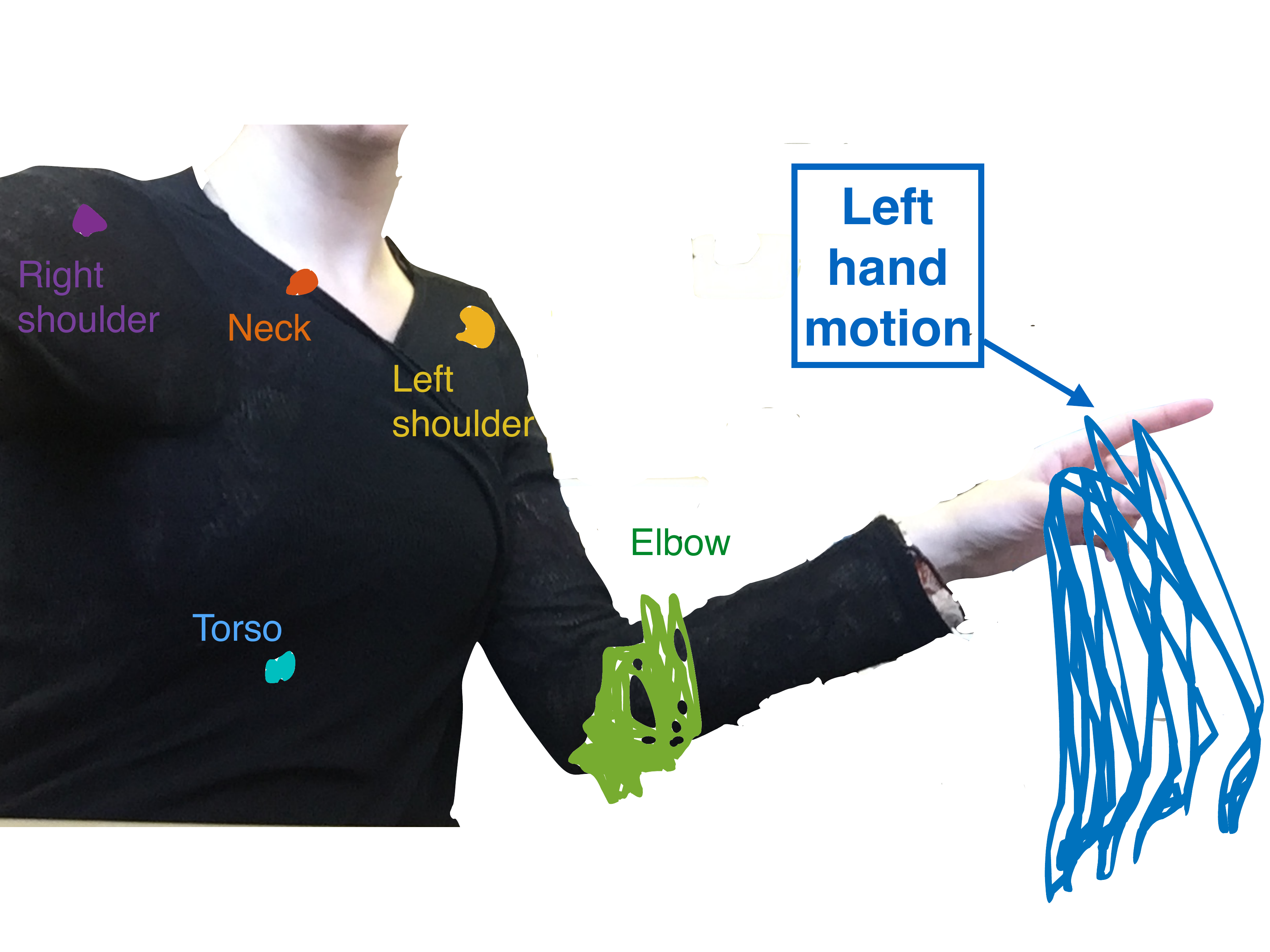}}
\hfill
\subfloat[Imperial-PRL KSC Dataset.]{\label{fig.maximekinectdata} \includegraphics[height=2.8cm]{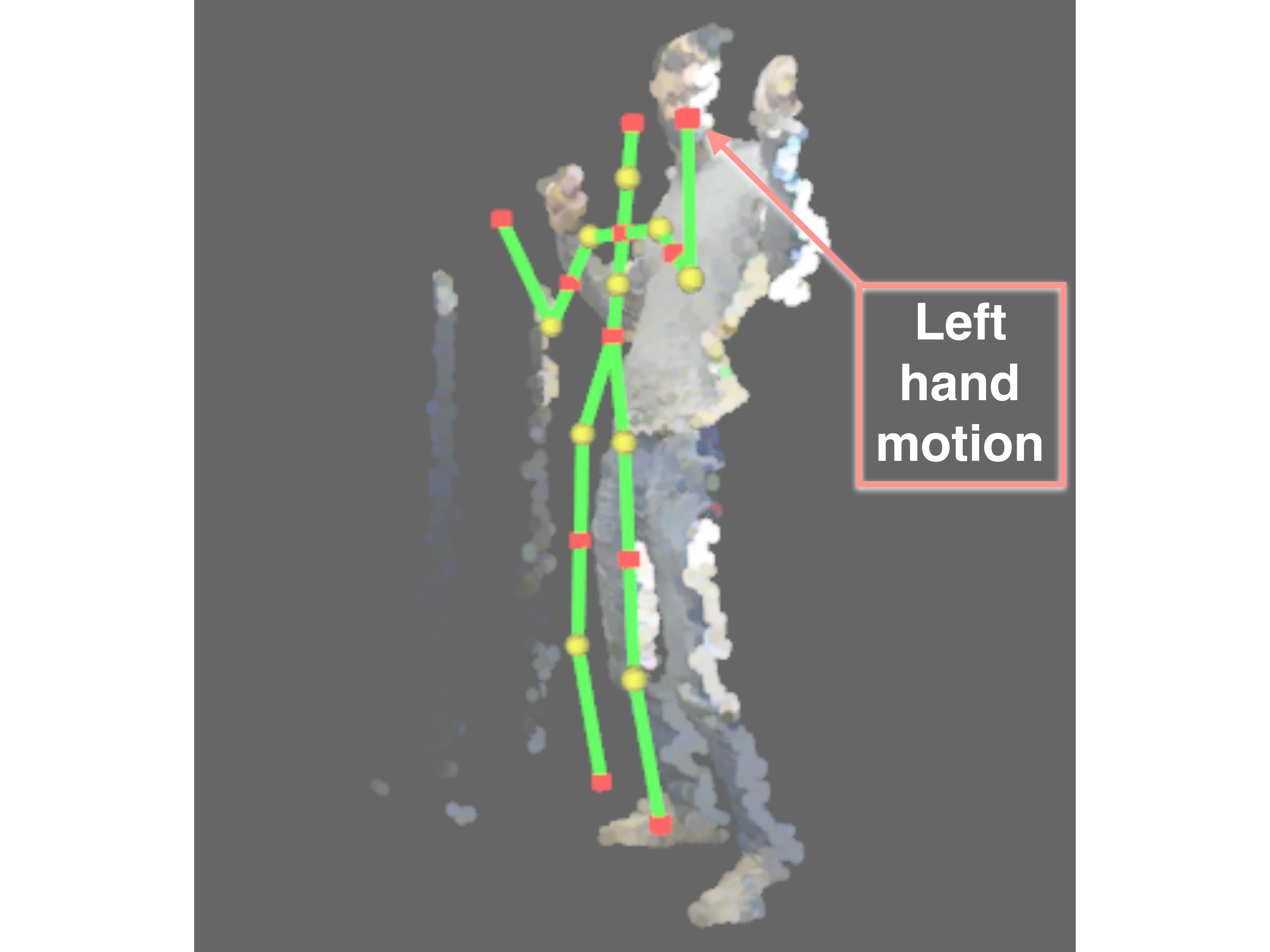}}  
\caption{
\protect\subref{fig.martinaPianokinectdata} Kinect data of a human upper-body movements while playing a piano keyboard with one hand.
\protect\subref{fig.maximekinectdata} Kinect data from the Imperial-PRL KSC Dataset. 
The trajectory of the left hand $\mathbf{v}_{\!_{\text{PRL}}}$ has been used as test dataset.
}
\end{figure}

Then we evaluated the architecture on data collected from the observation of other agents. 
Using the  multimodal variational autoencoder trained on data of the robot itself, the robot is able to make predictions also of others' motion trajectories in the visual space.
When observing others, the robot has access to the visual information only,  from its egocentric point of view. The learned model  
is then used to retrieve the motor commands (together with the other missing sensory modalities) that would enable the robot to reproduce the trajectory observed to perform mental simulation of the observed action.
Experiments were carried out using two different datasets. The first test dataset consists of movements of a human playing a piano keyboard, that was recorded by the authors using a RGB-D camera   (Fig.~\ref{fig.martinaPianokinectdata}). The second test dataset is part of the Imperial-PRL KSC Dataset (data used in \citep{chang2017learning} to validate kinematic structure correspondences methods). It contains kinect data of a human moving his hands (represented in Fig.~\ref{fig.maximekinectdata}). The 3D visual positions of these two datasets were then translated into 2D data by using two of the three available dimensions. This corresponds to a coarse approximation of the projection of the 3D trajectories onto the two cameras of the robot.

While the first dataset is similar to the self-exploration dataset in terms of scenario and application, the second one is significantly different, involving the free motion of the human arms, which are not confined within the scope of a keyboard.
The first test dataset allows us to demonstrate that the robot can effectively reconstruct and predict another agent's performing a sequence of motions that is similar to those performed in the motor babbling phase by using the learned internal models.
The second test dataset allows us to demonstrate that the robot is able to reconstruct and predict  visual trajectories of  others' motion using the learned models also when the type of motion is significantly different from the data acquired by the robot from self-exploration.

\begin{figure}
\includegraphics[width=\columnwidth]{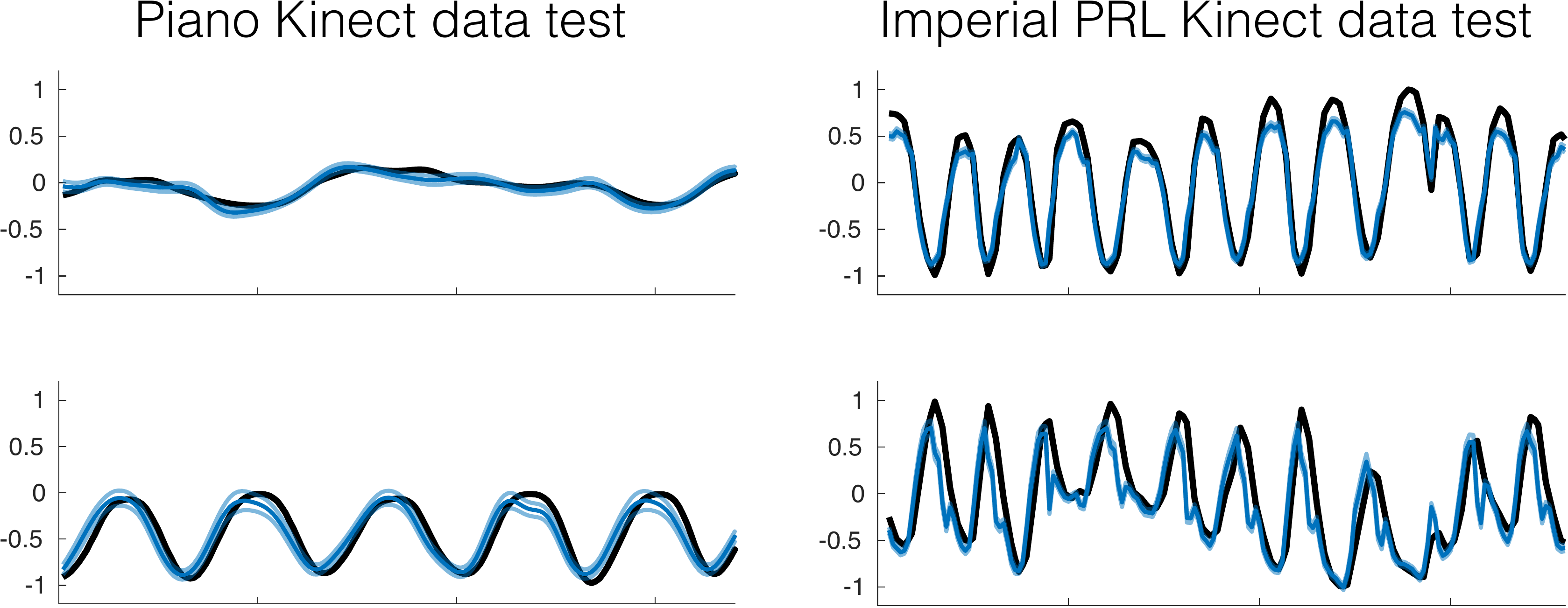}
\caption{Predictions of others' trajectories. Solid black lines represent the real data, while blue lines represent the predicted mean and the shaded light blue areas the predicted variance (uncertainty) of the prediction model. Prediction of human playing a piano keyboard (left) and prediction of the left hand motion $\mathbf{v}_{\!_{\text{PRL}}}$ (right).}
\label{fig.predother}
\end{figure}

Results are shown in Fig.~\ref{fig.predother}: the left plot shows the prediction performance on the kinect data collected from a human playing a piano keyboard (see Fig.~\ref{fig.martinaPianokinectdata}), and the right plot shows the prediction performance on the kinect data from the Imperial-PRL dataset (specifically on $\mathbf{v}_{\!_{\text{PRL}}}$, see Fig.~\ref{fig.maximekinectdata}).
The corresponding mean squared error scores obtained are $0.64\%$ and $0.69\%$ (corresponding to about 6 to 7 pixels) for the two datasets, respectively.  
The results achieved demonstrate that the proposed architecture obtains predictions of  visual trajectories of others' motion  by only making use of internal models of self.

\subsection{Imitate the observed agent's trajectories}
In this section, we present experiments that demonstrate the ability of the proposed architecture to use the learned multimodal representations to control the robot to imitate an observed agent 's visual trajectories. Condition (3) in Table~\ref{tab.training_struc} was critical to achieve this behavior. 
The experiments presented here show that the robot can successfully follow demonstrated/target visual trajectories, only using the learned multimodal representations. 

The learned model can  be used in a control loop (rightmost diagram in Fig.~\ref{fig.architecture}).
By deploying the learned model as a controller, it is possible to implement, for example, imitation tasks, where the robot needs to track trajectories in the sensory space. The learned model is able to reconstruct the motor commands necessary to achieve reference trajectories. The retrieved motor commands can then be issued to the robot's motors.
For this experiment, we have used two datasets: target trajectories from motor babbling, and data observed from the human playing two keys on the piano keyboard. 
The first dataset consists of trajectories from the part of the babbling dataset that has not been used for training the network. This test dataset thus contains data that have not been seen by the network before, though they are similar to the data used for training. In particular, the associations between positions in the sensory space and corresponding values of the velocity motor commands are similar.
The second dataset is more challenging, particularly because it may contain visual positions that were not contained in the training set, and this can in turn lead to combinations of the multimodal dimensions of the input that the network was never presented before.
The objective is for the robot to imitate the observed target trajectory. The target trajectory is used as reference and fed to the network in place of $\mathbf{v_t}$, while the current visual position of the robot and the current joint configuration of the robot ($\mathbf{v_{t-1}}$ and $\mathbf{q_{t-1}}$) are fed back to the network. All the other modalities are considered missing, in particular the motor commands that are produced by the network online after each new observation.

\begin{figure}
\includegraphics[width=\columnwidth]{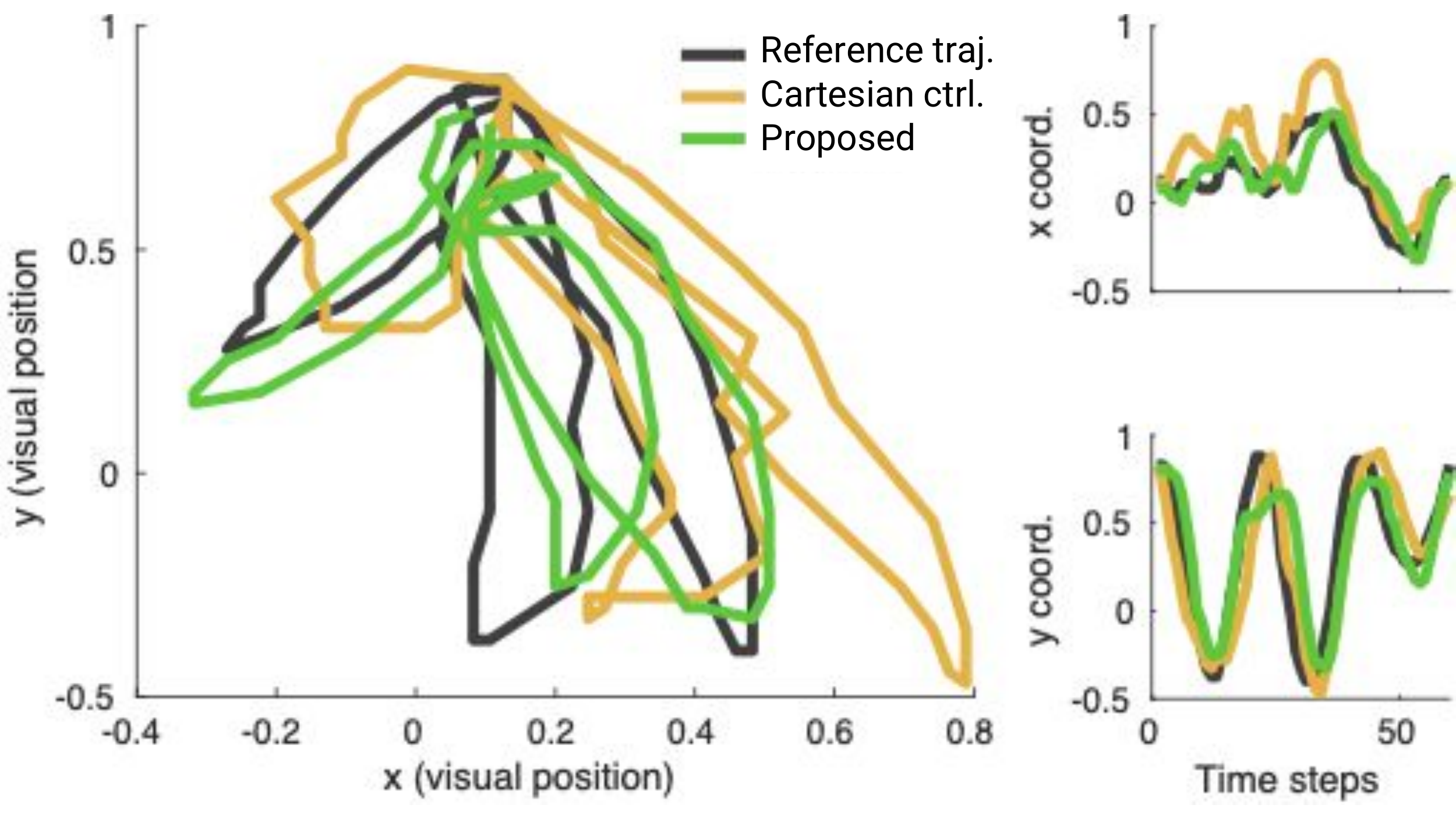}
\caption{Results of the imitation task realized by using the built-in Cartesian controller (yellow line) and the learned model (green line) to control online the robot's movements. The proposed method outperformed the built-in model, achieving a more accurate tracking of the reference visual trajectory (gray line).
The left plot shows the 2D visual position representation of the reference and executed trajectories, while the right plots show the corresponding temporal profiles of the positions ($x$ and $y$ coordinates). For  clarity of the representation, only the trajectories acquired from the left eye camera of the robot are depicted, while similar results were obtained from the right camera.
}
\label{fig.resultsctrl}
\end{figure}

In the first experiment, we have compared the proposed method with the Cartesian controller available on the iCub.
The stereo vision system of the iCub is used to determine the 3D position in the Cartesian space associated with 2D visual inputs. This information is then used by the Cartesian controller to reach the target positions.
Results obtained on the first dataset are represented in Fig.~\ref{fig.resultsctrl}.  The trajectories depicted in this figure are, consistently with the visual data used throughout this article, those captured from the robot's first person view. 
It can be noted that  our proposed method generates a trajectory that is more accurate than the one from  the built-in Cartesian controller. It is important to note  that the visual information available to the Cartesian controller is the same used by the proposed method, hence the calibration of the cameras together with the whole experimental setup is in common. This observation allows us to conclude that the proposed method overall surpasses the built-in controller in performing the task. 
The mean squared error score achieved by the proposed model on this task on the four-dimensional visual data is only  $0.48\%$ (corresponding to an error of less than 6 pixels), a very low value considering the resolution of the image ($320\times 240$ pixels)  and the precision of the visual data encoding the hand position throughout the experiments. 
The built-in Cartesian controller achieved a less accurate tracking of the reference visual trajectory, with a mean squared error score on the four-dimensional visual data of $1.07\%$, that is more than double the error achieved with the proposed method. 
 This difference is likely related to the fact that the reference data comes from the OpenCV tracker used to detect the 2D position of the hand in each image, which are probably not a perfect representation of the 3D position of the hand. This is likely causing the stereo-vision module to produce inaccurate target positions for the built-in Cartesian controller. Thus, we hypothesize that this succession of inaccuracies leads to a less accurate reproduction of the trajectory. Nevertheless, it is interesting to see that our proposed model manages to generate  a better trajectory, while using the same data and without the need of the prior knowledge contained in the built-in Cartesian controller.

\begin{figure}
\includegraphics[width=\columnwidth]{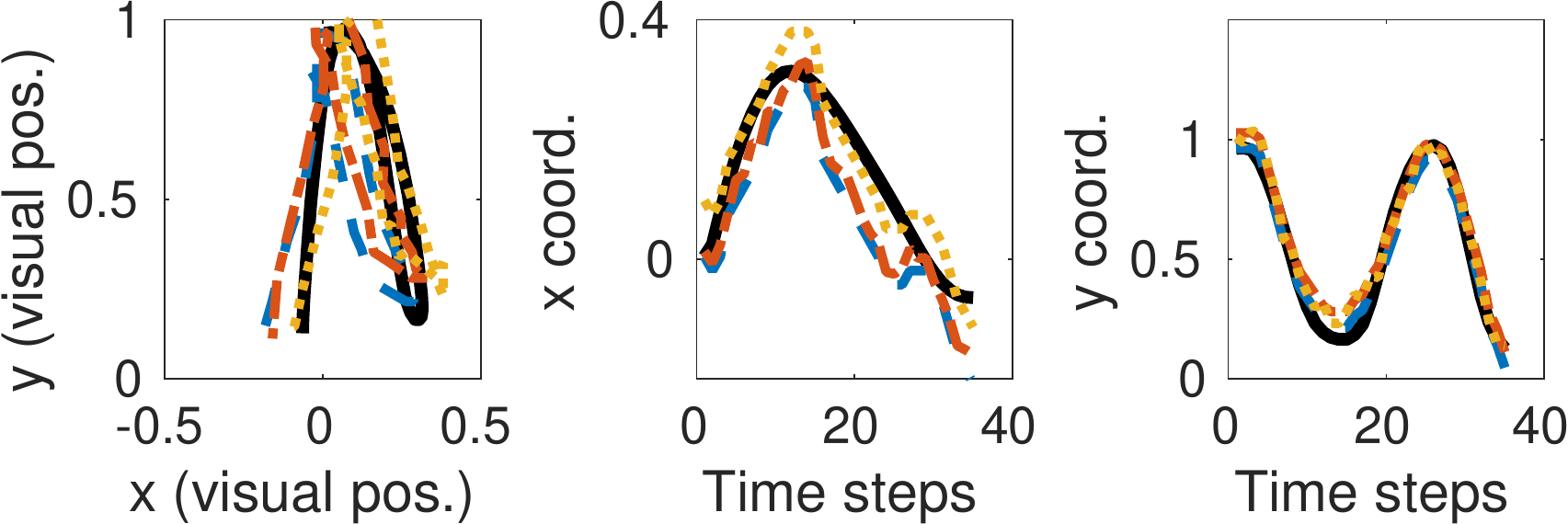}
\caption{Results of the imitation task on the data collected from a human playing a piano keyboard. The proposed method (colored lines) allows the robot to effectively track the reference visual trajectory (black line).
The left plot shows the 2D visual position representation of the reference and executed trajectories, while the right plots show the corresponding temporal profiles of the positions ($x$ and $y$ coordinates). For the clarity of the representation,  only 3 of the repetitions performed on the task are represented, and only the trajectories acquired from the left eye camera of the robot are depicted (analogous results where obtained from the right camera).
}
\label{fig.resultsctrl_piano}
\end{figure}

The experiments on the second dataset are also instrumental to show that the proposed method allows a robot to use data observed from another agent and imitate them. 
Results of 3 repetitions of this task are represented in Fig.~\ref{fig.resultsctrl_piano}. The mean squared error score achieved on this task on the four-dimensional visual data is $0.13\%$ (corresponding to an average of only 3 pixels error in the image frames).
The visual trajectory executed by the robot and represented in Fig.~\ref{fig.resultsctrl_piano} closely tracks the trajectory demonstrated. The robot is able to replicate the trajectory and  successfully hit the two keys that were played by the demonstrator. 
It is possible to note that the results on the $y$ coordinate are more accurate than those obtained on the $x$ coordinate. This reflects the structure of the actions performed during the exploration, which are used for training the model. While the exploratory movements spanned a wide range on the vertical direction, a smaller part of the space was explored on the horizontal direction.
We hypothesize that the bias observed in the network performance is related to the fact that the data acquired through the motor babbling exploration were also biased and constrained within a limited portion of the operational space. This limited, biased exploration allowed a more efficient data collection for the scope of the experiments and tasks described in this paper. 
We discuss this point further in Section~\ref{sec.discussion}.

\subsection{Results summary}

In summary, the proposed method achieves accurate reconstruction and prediction; moreover, it is able to generate control signals to imitate visual trajectories consistently and accurately.
We report in Table~\ref{tab.results_summary} a summary of the quantitative results obtained and described in the previous subsections.
The proposed method achieved low prediction errors across the different tasks considered: the model was able to predict with errors that can be considered negligible with respect to the state and action spaces (\textit{e.g.} less than 2 degrees angles for joint positions, less than 6 pixels in the vision space).

\begin{table}[]
\centering 
\caption{Accuracy scores summary: low prediction error is achieved on all the considered tasks, as only small discrepancies to the reference are measured.
\label{tab.results_summary}}
\resizebox{\columnwidth}{!}{%
\begin{tabular}{@{}ll@{}}
\toprule
\multicolumn{1}{l|}{Task} & \begin{tabular}[c]{@{}l@{}}Accuracy \\ (percentage scores, relative to the dataset ranges)\end{tabular} \\ \midrule
\multicolumn{1}{l|}{Reconstruction} & \begin{tabular}[c]{@{}l@{}}Joint pos.: 0.46\% ($\approx 1.29$ degrees)\\ Vision: 0.05\% ($\approx 1.85$ pixels)\\ Touch: 2.35\% \\ Sound: 3.35\%\\ Motor c.: 1.29\% ($\approx 2.16$ degree per sec.)\end{tabular} \\\midrule
\multicolumn{1}{l|}{\begin{tabular}[c]{@{}l@{}}Reconstruction\\ from partial data\end{tabular}} & \begin{tabular}[c]{@{}l@{}}Joint pos.: 1.39\% ($\approx 2.24$ degrees)\\ Vision: 0.05\% ($\approx 1.85$ pixels)\\ Touch: 9.42\%\\ Sound: 3.95\%\\ Motor c.: 2.32\% ($\approx 2.89$ degrees per sec.)\end{tabular} \\ \midrule
\multicolumn{1}{l|}{\begin{tabular}[c]{@{}l@{}}Prediction\\ of self motion\end{tabular}} & \begin{tabular}[c]{@{}l@{}}Single step: 0.21\% ($<4$ pixels)\\ Multi-step: 0.42\% ($<5$ pixels, after 50 time steps)\end{tabular}

 \\ \midrule
\multicolumn{1}{l|}{\begin{tabular}[c]{@{}l@{}}Prediction\\ of others motion\end{tabular}} & \begin{tabular}[c]{@{}l@{}}Piano playing: 0.64\% ($\approx 6$ pixels)\\ Imperial-PRL-KSC: 0.69\% ($\approx 6$ pixels)\end{tabular} \\ \midrule
\multicolumn{1}{l|}{Imitation} & \begin{tabular}[c]{@{}l@{}}Motor babbling: 0.48\% ($\approx 5$ pixels)\\ Playing keys: 0.13\% ($\approx 3$ pixels)\end{tabular}  \\ \bottomrule
\end{tabular}
}
\end{table}

\subsection{Comparison with other methods}

An important aspect of this work is that the training procedure can be applied to other neural network architectures with reconstruction capabilities. 
The augmentation of the dataset with different arrangements of missing modalities enables the construction of a single model capable of executing several tasks. In order to illustrate the possibility of applying this training procedure to other networks, and to compared the accuracy of the proposed model, we have tested several other architectures:
\begin{itemize}
\item[(i)] Vanilla VAE: a standard VAE model (\textit{e.g.}~\citep{kingma2013auto}) trained in a denoising fashion on the dataset without missing modalities, by using a probability of 30\% to set some values of the inputs to $0$.
\item[(ii)] Vanilla VAE trained with our proposed training method, on the augmented database. 
\item[(iii)] The multimodal architecture proposed in \citep{Droniou2015}. This architecture learns a shared latent representation and classification of the inputs, and can be used to reconstruct missing modalities. It is trained as (i), which is also the approach proposed by the authors. The implementation of this architecture is based on the source code provided by its authors. The sizes of the different fully connected layers have been selected to match those of our proposed architecture.
\item[(iv)] The multimodal architecture proposed in \citep{Droniou2015} trained with our augmented database. 
\item[(v)] Two independent models, namely a forward and an inverse models, implemented by feed-forward neural networks.
\end{itemize}

\begin{figure}
\centering
\includegraphics[width=\columnwidth]{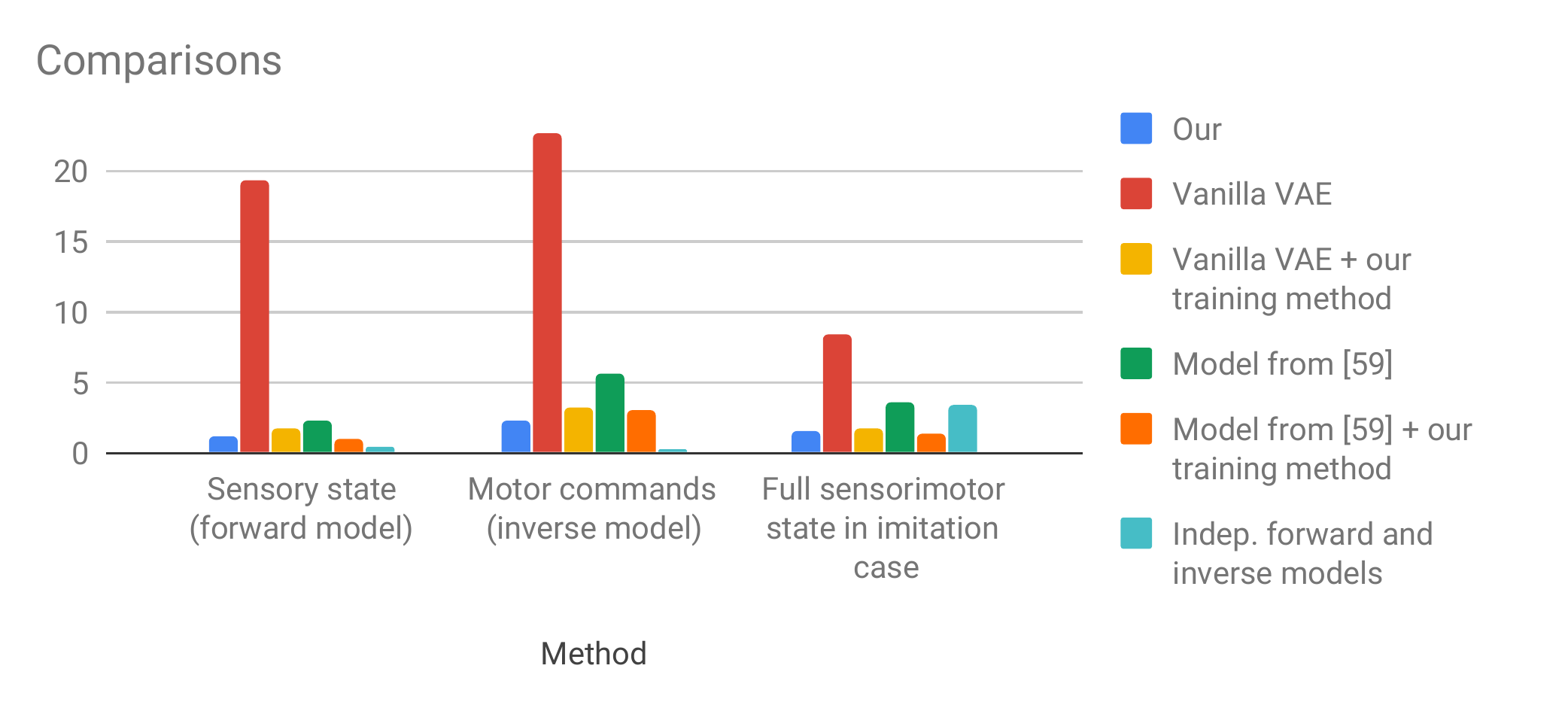}
\caption{Visualization of performance scores (prediction error) of the proposed method and other methods. Our training strategy clearly improves performance of reconstruction methods, including Vanilla VAE and the model proposed in \citep{Droniou2015}. Our method performs equally or better than the alternatives in the complex fully sensorimotor state estimation task.}
\label{fig.chart}
\end{figure}

Implementation details of the different architectures are given in~\ref{app.architecture}.	
We considered three representative cases for comparison, namely:
\begin{itemize}
\item prediction of the current sensory state from the previous one (case 2 of Table~\ref{tab.training_struc}); this case corresponds to the forward model function;
\item prediction of the motor commands from the visual information only (case 4 of Table~\ref{tab.training_struc}); this case corresponds to the inverse model function;
\item prediction of the whole sensorimotor state from the external visual information and the current joints configuration (case 3 of Table~\ref{tab.training_struc}); this case corresponds to the imitation scenario. 
\end{itemize} 
Fig.~\ref{fig.chart} provides a visual representation of the performance comparisons in terms of prediction error.
Table~\ref{tab.rec_comparison} summarizes the MSE scores obtained by the models compared.
From the results presented in Table~\ref{tab.rec_comparison}, we are able to draw the following conclusions.
First, the proposed training strategy consistently improves the performance of the considered models, allowing a drop of the MSE scores to approximately half of the original scores in the case of the model from \citep{Droniou2015}, and to a fraction of it in the case of the vanilla VAE. Also, the proposed multimodal VAE outperforms a vanilla VAE model: we argue that this is because the proposed multimodal model can learn both modality-specific and cross-modality features thanks to the modular structure of the encoder/decoder and the joint probability distribution learned in the latent encoding.

The comparison with the two independent forward and inverse models demonstrates that the proposed architecture performs better because it can fulfill the two functions (of forward and inverse model) simultaneously.
In this comparison, the predictions from the forward model are used for the ``forward model'' case, and the prediction from the inverse model are used for the ``inverse model'' case, respectively. 
To achieve the third case (imitation case) the forward and inverse models must feed each other in order to produce the whole sensorimotor state from the visual and proprioception information: first, the inverse model must be applied to get the motor commands which are then used by the forward model to produce the sensory state prediction. Despite each individual model being (almost) perfectly suited for its own function (note the lowest scores achieved), the combination of the two to achieve imitation results does not achieve the best performance on the imitation case.
On the contrary, the proposed architecture outperforms this baseline.

\begin{table}
\centering 
\caption{ Accuracy of different architectures on the tasks presented in this paper. The training and evaluation of the different models have been replicated 10 times. The results are presented in the form of percentages indicating: median [first quartile; third quartile].
\label{tab.rec_comparison}}
\resizebox{\columnwidth}{!}{%
\begin{tabular}{l | c c c}
\hline\hline
Method &
\begin{tabular}[c]{@{}c@{}}Prediction of \\ sensory state \\ (\textit{forward model})\end{tabular} &
\begin{tabular}[c]{@{}c@{}}Prediction of \\ motor command \\ (\textit{inverse model})\end{tabular} &
\begin{tabular}[c]{@{}c@{}}Prediction of \\ full sensorimotor state \\ in imitation case\end{tabular} 
\\ [0.5ex] \hline\hline \\ [-1ex] 

Our	&\textbf{1.13\% [0.96; 1.22]\%} & \textbf{2.31\% [2.28; 2.34]\%} & \textbf{1.52\% [1.45; 1.67]\%}   
\\  \hline 
\\[-1ex]

Vanilla VAE & 19.37\% [12.48; 35.39]\% & 22.65\% [12.47; 49.08]\%         & 8.43\% [6.86; 9.16]\% \\ \hline 
\\[-1ex]

\begin{tabular}[l]{@{}l@{}}Vanilla VAE \\ $\quad$+ our training method 	\end{tabular}		& 1.75\% [1.63; 1.82]\% & 3.31\% [3.24; 3.73]\%         & 1.76\% [1.72; 1.81]\%    
\\ \hline 
\\[-1ex]

Model from \citep{Droniou2015} 							& 2.36\% [2.07;2.61]\% & 5.72\% [5.66; 5.77]\%         & 3.56\% [3.22; 3.71]\%      
\\ \hline \\[-1ex]

\begin{tabular}[l]{@{}l@{}}Model from \citep{Droniou2015}  \\ $\quad$+ our training method 	\end{tabular}	& \textbf{1.09\% [1.06; 1.12]\%} & 3.03\% [2.91; 3.06]\% & \textbf{1.45\% [1.43; 1.47]\%}      
\\ \hline 
\\[-1ex]

\begin{tabular}[l]{@{}l@{}}Indep. forward \\ $\quad$ and inverse models 	\end{tabular}			& 0.51\% [0.49; 0.53]\% & 0.24\% [0.23; 0.26]\%         & 3.39\% [3.04; 3.64]\% 
\\ [0.5ex] \hline 
\end{tabular}
} 
\end{table}

\section{Discussion}
\label{sec.discussion}

The results presented in this study show that a robot can learn to predict  the visual trajectories of another agent from an egocentric point of view by exploiting only self-learned internal models.
In this study, it has been argued that one of the main challenges in achieving predictions of others only based on internal models of self is the difference of the available data:
while the whole set of sensorimotor data is available when the robot is acting and exploring, only visual information is available when the robot observes another agent.
This motivated the proposed strategy to reconstruct and infer the missing information. 
In particular,  the proposed training strategy has shown crucial to improve models performance, and 
the proposed variational autoencoder allowed a robot to learn probability distributions among different sensorimotor modalities which captures the kinematic redundancy  of the robot's motions. 

The choice of the variational autoencoder was motivated by its capability of modelling data uncertainty, through a learned posterior distribution represented by the mean and the variance of a Gaussian distribution. 
The multimodal formulation, moreover, allows us to combine different representations of different types of data into a single distribution (the learned posterior distribution), that gracefully merges the different sources of information. 
In addition, the encoder-decoder structure of the variational autoencoder is ideal for reconstruction and self-supervised learning purposes, hence a perfect fit for the objective of this work: that is to reproduce (reconstruct, predict, generate) signals during inference, after training on exploration (self-collected) data.
The choice of a variational autoencoder model instead of a classical autoencoder also allowed us to leverage the advantages of generative models.
Variational autoencoders model the input data by means of a distribution, generally (as in our case) a Gaussian distribution, defined by a mean and a variance. This allows to capture a more general and flexible underlying structure of the data compared to other models (such as standard autoencoders or encoder-decoder models). In our case, the distribution is action-conditioned since part of the input includes the motor commands. This means that the posterior distribution learned during training captures the correspondences between actions and sensor observations, and learns that some observations actually correspond to different actions. This is shown in Figure~\ref{fig.vae-redundancy}: despite the fact that joint $q_2$ does not follow the prescribed trajectory, the visual trajectory (as well as tactile and acoustic ones) is actually tracked accurately. This is because the same visual position of the hand can be achieved by a number of different joint configurations (redundancy). The fact that the variance of joint $q_2$ is significantly bigger than the variance of the other joints supports this claim, because it represents the uncertainty of this particular joint motion.

The proposed approach can be enhanced by 
enforcing the variational autoencoder to learn a latent space of a certain shape from which inputs can be sampled in a more meaningful manner, to generate synthetic sensorimotor data. 
Although we let the exploration of this direction for future work, we believe this is a strong and promising characteristic of the chosen model in the context of multimodal learning.
Another key characteristic of the proposed multimodal variational autoencoder is that this model can learn both modality-specific and cross-modality features thanks to the modular structure of the encoder/decoder and the joint probability distribution learned in the latent encoding.

A limitation of the current implementation is the dependence of the reconstruction accuracy on the explored sensorimotor space. In particular, it is possible that combinations of sensory states reached during an imitation task are far from the training set of states used in the training of the network. In this case the network ``guesses'' motor commands by sampling from the learned distribution, but the reconstruction accuracy is usually poor due to the lack of samples resembling the observed new sensory state.

The problem of generalizing to unexplored regions of the space is indeed a very interesting and still largely unsolved problem in robotics as well as in exploration methods in other domains (\textit{e.g.} machine learning, reinforcement learning, multi-task learning, \textit{etc.}). One possibility to improve our current method would be to enlarge the exploration space to include a larger region of the multimodal space (\textit{e.g.} bigger areas of Cartesian/join space). This would come with the problem of having to acquire larger number of data and thus making learning of the model slower.
A possible direction is the implementation of more sophisticated exploration strategies, for instance curiosity-based strategies \citep{maestre2015bootstrapping,baranes2010intrinsically}, or to exploit the generative nature of the model as mentioned earlier.

Finally, in this paper, we designed the tasks in a way that the robot and the human are both capable of executing it. It would be interesting in future works to investigate how to identify and address the situation when the task cannot be fulfilled by the robot.

\section{Conclusion and Future Work}

This work takes inspiration from cognitive studies showing that humans can predict others' actions by using their own internal models \citep{demiris2014mirror}. Following this direction, we have implemented a new architecture that allows a robot to predict  visual trajectories of other agents' actions by using only self-learned internal models. 
In this paper, we introduced a  strategic training approach and a  multimodal learning architecture that allow a robot to (1) reconstruct missing sensory modalities, (2) predict the  its own sensorimotor state and predict visual trajectories of another agent from an egocentric point of view, and (3) imitate the observed agent 's trajectories.
This versatility represents a major advantage of the proposed approach, that can thus be applied in different applications to address different objectives (\textit{e.g.} prediction, control, etc.).
This architecture leverages advantages of developmental robotics and of deep learning,
and has been evaluated extensively on different datasets and set-ups.

In future work, we will investigate how to leverage the generative capabilities of the network, and how this method can be combined with more advanced exploration strategies (such as curiosity-based strategies) in order to acquire a self-perception database that covers the robot and environment states as much as possible \citep{maestre2015bootstrapping,baranes2010intrinsically}. The presented method will also be combined with perspective taking mechanisms \citep{JohnsonDemiris05, Fischer2016} to enable prediction of future states from different viewpoints.

\section*{Acknowledgements}

This work was supported by an EPSRC doctoral scholarship (Grant Number 1507722), EU FP7 project WYSIWYD under Grant 612139, and EU Horizon2020 project PAL under Grant 643783-RIA.

\appendix
\section{Datasets and training}
\label{app.datatrain}

The motor babbling dataset contains 7380 datapoints (corresponding to approximately 30 minutes of exploration). The trajectory taken from the Imperial-PRL KSC dataset contains 25 datapoints (corresponding to approximately 45 seconds).
The VAE is trained on the motor babbling data augmented training set, for 80000 epochs, with learning rate of 0.00005, and a batch size of 1000 samples. At each training step, a batch is randomly sampled from the augmented training set and fed to the network to train. 
The augmented training set is formed by concatenating the original complete set of data collected during motor babbling and normalized to values between -1 and 1, with mutilated versions of it: Table 1 shows how the augmented dataset is formed: (1) complete data at time $t-1$ and $t$, concatenated to (2) data including only time $t-1$, concatenated to (3) data including only proprioception at time $t-1$ and vision at time $t$ and $t-1$, concatenated to (4) data including only vision at $t$ and $t-1$. For the cases (2-3-4), the missing data is replaced with the value $-2$ (which is outside of the normalized range $[-1,1]$ used for the collected data).
Each dataset was split with a 80:20 ratio between training and testing datapoints. Given the size of the dataset, the model can  overfit to the training set. Nonetheless, because the training dataset was collected by using pseudo-random movements (\textit{i.e.} not specific to a particular task to be performed), the network is able to generalize to different types of motion.

\section{Parameters of the architectures}
\label{app.architecture}

\noindent
Multimodal Variational Autoencoder (proposed architecture)
 
\noindent\resizebox{\columnwidth}{!}{%
\begin{tabular}{|c|c|c|c|c|l}
\cline{1-5}
joint positions& visual & tactile & sound& motor commands& input layer\\
8 dims&8 dims&2 dims&2 dims&8 dims\\
\cline{1-5}
40-ReLU& 40-ReLU& 10-ReLU& 10-ReLU & 40-ReLU& Modality encoders\\
\cline{1-5}
20-ReLU& 20-ReLU& 5-ReLU& 5-ReLU & 20-ReLU\\
\cline{1-5}
\multicolumn{5}{|c|}{concatenation}\\
\cline{1-5}
\multicolumn{5}{|c|}{100-ReLU}& Shared encoder\\
\cline{1-5}
\multicolumn{5}{|c|}{28-ReLU x2}& Latent space\\
\cline{1-5}
\multicolumn{5}{|c|}{100-ReLU}& Shared decoder\\
\cline{1-5}
\multicolumn{5}{|c|}{70-ReLU}\\
\cline{1-5}
\multicolumn{5}{|c|}{slicing into 20, 20, 5, 5, 20 dimensions respectively}\\
\cline{1-5}
40-ReLU& 40-ReLU& 10-ReLU& 10-ReLU & 40-ReLU &Modality decoders\\
\cline{1-5}
8-ReLU x2& 8-ReLU x2& 2-ReLU x2& 2-ReLU x2 & 8-ReLU x2& Reconstructed data\\
\cline{1-5}
\end{tabular}
}
\newline

\noindent
N-ReLU represents a fully connected layer with N neurons and using the ReLU activation function. N-ReLU x2 indicates that 2 N-ReLU layers are created in parallel, one to encode the mean and the other to encode the variance of the output distribution. The network has been trained for 80k epochs with the Adam optimizer and a learning rate of $0.00005$. The training took approximately 5 hours on a single GPU (Nvidia GTX-1080).

Structure of the compared approaches:

\noindent\resizebox{\columnwidth}{!}{%
\begin{tabular}{|c|l|c|l|c|l}
\multicolumn{2}{l}{VAE}& \multicolumn{2}{l}{Forward Model} & \multicolumn{2}{l}{Inverse Model} \\
\cline{1-1}\cline{3-3}\cline{5-5}
 all modalities& input layer&		all modalities at t-1 & input layer&	Sensory state at t-1 and t& input layer\\
 28 dims&&							14 dims&&								20 dims\\
 \cline{1-1}\cline{3-3}\cline{5-5}
 100-ReLU& Encoders&				14-tanh&&								100-tanh& \\
\cline{1-1}\cline{3-3}\cline{5-5}
 100-ReLU&&							10-linear& output layer&				100-tanh& \\
\cline{1-1}\cline{3-3}\cline{5-5}
 28-ReLU x2& \multicolumn{2}{|l}{Latent space}		&&										4-linear& output layer\\
\cline{1-1}\cline{5-5}
 100-ReLU& \multicolumn{2}{|l}{Encoders}				\\
\cline{1-1}
 100-ReLU\\
\cline{1-1}
 28-ReLU x2& \multicolumn{2}{|l}{Reconstructed data}\\
\cline{1-1}
\end{tabular}
}\newline

\noindent
The implementation of the comparison architecture from \citep{Droniou2015} is based on the source code provided by the authors and replicate most of its parameters. Only differences are the number of modalities (set to 5), the number of parameters (set to 100), and the number of classes (set to one as classification is not considered here). 


\bibliographystyle{elsarticle-num} 
\bibliography{references}

\begin{thebibliography}{10}
\expandafter\ifx\csname url\endcsname\relax
  \def\url#1{\texttt{#1}}\fi
\expandafter\ifx\csname urlprefix\endcsname\relax\def\urlprefix{URL }\fi
\expandafter\ifx\csname href\endcsname\relax
  \def\href#1#2{#2} \def\path#1{#1}\fi

\bibitem{Wolpert2001}
D.~M. Wolpert, J.~R. Flanagan, {Motor prediction.}, Current biology 11~(18)
  (2001) R729--R732 (2001).

\bibitem{Wolpert1998}
D.~Wolpert, M.~Kawato, {Multiple paired forward and inverse models for motor
  control}, Neural Networks 11~(7) (1998) 1317--1329 (1998).

\bibitem{Kawato1999}
M.~Kawato, {Internal models for motor control and trajectory planning}, Current
  Opinion in Neurobiology 9~(6) (1999) 718--727 (1999).

\bibitem{demiris2006hierarchical}
Y.~Demiris, B.~Khadhouri, Hierarchical attentive multiple models for execution
  and recognition of actions, Robotics and autonomous systems 54~(5) (2006)
  361--369 (2006).

\bibitem{demiris2014mirror}
Y.~Demiris, L.~Aziz-Zadeh, J.~Bonaiuto, Information processing in the mirror
  neuron system in primates and machines, Neuroinformatics 12~(1) (2014) 63--91
  (2014).

\bibitem{hafner2005interpersonal}
V.~Hafner, F.~Kaplan, Interpersonal maps and the body correspondence problem,
  in: Proceedings of the Third International Symposium on Imitation in animals
  and artifacts, Citeseer, 2005, pp. 48--53 (2005).

\bibitem{alissandrakis2002imitation}
A.~Alissandrakis, C.~L. Nehaniv, K.~Dautenhahn, Imitation with alice: Learning
  to imitate corresponding actions across dissimilar embodiments, IEEE
  Transactions on Systems, Man, and Cybernetics-Part A: Systems and Humans
  32~(4) (2002) 482--496 (2002).

\bibitem{nehaniv1998mapping}
C.~Nehaniv, K.~Dautenhahn, Mapping between dissimilar bodies: A ordances and
  the algebraic foundations of imitation, EWLR-98 (1998) 64--72 (1998).

\bibitem{JohnsonDemiris05}
M.~Johnson, Y.~Demiris, Perceptual perspective taking and action recognition,
  International Journal of Advanced Robotic Systems 2~(4) (2005) 301--308
  (2005).

\bibitem{Fischer2016}
T.~Fischer, Y.~Demiris, {Markerless Perspective Taking for Humanoid Robots in
  Unconstrained Environments}, in: IEEE International Conference on Robotics
  and Automation, 2016, pp. 3309--3316 (2016).

\bibitem{baraglia2015motor}
J.~Baraglia, J.~L. Copete, Y.~Nagai, M.~Asada, Motor experience alters action
  perception through predictive learning of sensorimotor information, in: Joint
  IEEE International Conference on Development and Learning and Epigenetic
  Robotics, IEEE, 2015, pp. 63--69 (2015).

\bibitem{copete2016motor}
J.~L. Copete, Y.~Nagai, M.~Asada, Motor development facilitates the prediction
  of others' actions through sensorimotor predictive learning, in: Joint IEEE
  International Conference on Development and Learning and Epigenetic Robotics,
  IEEE, 2016, pp. 223--229 (2016).

\bibitem{kamel2018deep}
A.~Kamel, B.~Sheng, P.~Yang, P.~Li, R.~Shen, D.~D. Feng, Deep convolutional
  neural networks for human action recognition using depth maps and postures,
  IEEE Transactions on Systems, Man, and Cybernetics: Systems (2018).

\bibitem{kamel2019efficient}
A.~Kamel, B.~Sheng, P.~Li, J.~Kim, D.~D. Feng, Efficient body motion
  quantification and similarity evaluation using 3d joints skeleton
  coordinates, IEEE Transactions on Systems, Man, and Cybernetics: Systems
  (2019).

\bibitem{kamel2019investigation}
A.~Kamel, B.~Liu, P.~Li, B.~Sheng, An investigation of 3d human pose estimation
  for learning tai chi: A human factor perspective, International Journal of
  Human-Computer Interaction 35~(4-5) (2019) 427--439 (2019).

\bibitem{cully2015robots}
A.~Cully, J.~Clune, D.~Tarapore, J.-B. Mouret, Robots that can adapt like
  animals, Nature 521~(7553) (2015) 503--507 (2015).

\bibitem{kriegman2019automated}
S.~Kriegman, S.~Walker, D.~Shah, M.~Levin, R.~Kramer-Bottiglio, J.~Bongard,
  Automated shapeshifting for function recovery in damaged robots, in:
  Proceedings of Robotics: Science and System XV (RSS), 2019 (2019).

\bibitem{kingma2013auto}
D.~P. Kingma, M.~Welling, Auto-encoding variational bayes, arXiv preprint
  arXiv:1312.6114 (2013).

\bibitem{rezende2014stochastic}
D.~J. Rezende, S.~Mohamed, D.~Wierstra, Stochastic backpropagation and
  approximate inference in deep generative models, arXiv preprint
  arXiv:1401.4082 (2014).

\bibitem{sutton1998reinforcement}
R.~S. Sutton, A.~G. Barto, Reinforcement learning: An introduction, Vol.~1, MIT
  press Cambridge, 1998 (1998).

\bibitem{abbeel2007application}
P.~Abbeel, A.~Coates, M.~Quigley, A.~Y. Ng, An application of reinforcement
  learning to aerobatic helicopter flight, Advances in neural information
  processing systems 19 (2007) 1 (2007).

\bibitem{argall2009survey}
B.~D. Argall, S.~Chernova, M.~Veloso, B.~Browning, A survey of robot learning
  from demonstration, Robotics and autonomous systems 57~(5) (2009) 469--483
  (2009).

\bibitem{billard2008robot}
A.~Billard, S.~Calinon, R.~Dillmann, S.~Schaal, Robot programming by
  demonstration, in: Springer handbook of robotics, Springer, 2008, pp.
  1371--1394 (2008).

\bibitem{deisenroth2011pilco}
M.~Deisenroth, C.~E. Rasmussen, {PILCO: A model-based and data-efficient
  approach to policy search}, in: Proceedings of the 28th International
  Conference on machine learning, 2011, pp. 465--472 (2011).

\bibitem{williams2009multi}
C.~Williams, S.~Klanke, S.~Vijayakumar, K.~M. Chai, Multi-task gaussian process
  learning of robot inverse dynamics, in: Advances in Neural Information
  Processing Systems, 2009, pp. 265--272 (2009).

\bibitem{miller1995neural}
W.~T. Miller, P.~J. Werbos, R.~S. Sutton, Neural networks for control, MIT
  press, 1995 (1995).

\bibitem{kawato1988hierarchical}
M.~Kawato, Y.~Uno, M.~Isobe, R.~Suzuki, Hierarchical neural network model for
  voluntary movement with application to robotics, IEEE Control Systems
  Magazine 8~(2) (1988) 8--15 (1988).

\bibitem{Hinton2006}
G.~E. Hinton, S.~Osindero, Y.-W. Teh, {A Fast Learning Algorithm for Deep
  Belief Nets}, Neural Computation 18~(7) (2006) 1527--1554 (2006).

\bibitem{levine2016end}
S.~Levine, C.~Finn, T.~Darrell, P.~Abbeel, End-to-end training of deep
  visuomotor policies, Journal of Machine Learning Research 17~(39) (2016)
  1--40 (2016).

\bibitem{Lungarella2003}
M.~Lungarella, G.~Metta, R.~Pfeifer, G.~Sandini, {Developmental robotics: a
  survey}, Connection Science 15~(4) (2003) 151--190 (2003).

\bibitem{Sigaud2016}
O.~Sigaud, A.~Droniou, {Towards Deep Developmental Learning}, IEEE Transactions
  on Cognitive and Developmental Systems 8~(2) (2016) 99--114 (2016).

\bibitem{droniou2012autonomous}
A.~Droniou, S.~Ivaldi, V.~Padois, O.~Sigaud, Autonomous online learning of
  velocity kinematics on the icub: A comparative study, in: IEEE/RSJ
  International Conference on Intelligent Robots and Systems, IEEE, 2012, pp.
  3577--3582 (2012).

\bibitem{vicente2016online}
P.~Vicente, L.~Jamone, A.~Bernardino, Online body schema adaptation based on
  internal mental simulation and multisensory feedback, Frontiers in Robotics
  and AI 3 (2016) 7 (2016).

\bibitem{calandra2015learning}
R.~Calandra, S.~Ivaldi, M.~P. Deisenroth, E.~Rueckert, J.~Peters, Learning
  inverse dynamics models with contacts, in: IEEE International Conference on
  Robotics and Automation, IEEE, 2015, pp. 3186--3191 (2015).

\bibitem{fitzpatrick2006reinforcing}
P.~Fitzpatrick, A.~Arsenio, E.~R. Torres-Jara, Reinforcing robot perception of
  multi-modal events through repetition and redundancy and repetition and
  redundancy, Interaction Studies 7~(2) (2006) 171--196 (2006).

\bibitem{ruesch2008multimodal}
J.~Ruesch, M.~Lopes, A.~Bernardino, J.~Hornstein, J.~Santos-Victor, R.~Pfeifer,
  Multimodal saliency-based bottom-up attention a framework for the humanoid
  robot icub, in: IEEE International Conference on Robotics and Automation,
  IEEE, 2008, pp. 962--967 (2008).

\bibitem{Demiris2005crib}
Y.~Demiris, A.~Dearden, {From motor babbling to hierarchical learning by
  imitation: a robot developmental pathway}, International Workshop on
  Epigenetic Robotics: Modeling Cognitive Development in Robotic Systems (2005)
  31--37 (2005).

\bibitem{schaal2003computational}
S.~Schaal, A.~Ijspeert, A.~Billard, Computational approaches to motor learning
  by imitation, Philosophical Transactions of the Royal Society B: Biological
  Sciences 358~(1431) (2003) 537--547 (2003).

\bibitem{calinon2010probabilistic}
S.~Calinon, F.~D'halluin, E.~Sauser, D.~Caldwell, A.~Billard, A probabilistic
  approach based on dynamical systems to learn and reproduce gestures by
  imitation, IEEE Robotics and Automation Magazine 17~(2) (2010) 44--54 (2010).

\bibitem{lopes2005visual}
M.~Lopes, J.~Santos-Victor, Visual learning by imitation with motor
  representations, IEEE Transactions on Systems, Man, and Cybernetics, Part B
  (Cybernetics) 35~(3) (2005) 438--449 (2005).

\bibitem{nehaniv2001like}
C.~L. Nehaniv, K.~Dautenhahn, Like me?-measures of correspondence and
  imitation, Cybernetics \& Systems 32~(1-2) (2001) 11--51 (2001).

\bibitem{shimizu2014robust}
T.~Shimizu, R.~Saegusa, S.~Ikemoto, H.~Ishiguro, G.~Metta, Robust sensorimotor
  representation to physical interaction changes in humanoid motion learning,
  IEEE transactions on neural networks and learning systems 26~(5) (2014)
  1035--1047 (2014).

\bibitem{ramisa2017breakingnews}
A.~Ramisa, F.~Yan, F.~Moreno-Noguer, K.~Mikolajczyk, Breakingnews: Article
  annotation by image and text processing, IEEE Transactions on pattern
  analysis and machine intelligence (2017).

\bibitem{poria2016fusing}
S.~Poria, E.~Cambria, N.~Howard, G.-B. Huang, A.~Hussain, Fusing audio, visual
  and textual clues for sentiment analysis from multimodal content,
  Neurocomputing 174 (2016) 50--59 (2016).

\bibitem{Ngiam2011}
J.~Ngiam, A.~Khosla, M.~Kim, J.~Nam, H.~Lee, A.~Y. Ng, {Multimodal Deep
  Learning}, Proceedings of The 28th International Conference on Machine
  Learning (2011) 689--696 (2011).

\bibitem{zambelli2016}
M.~Zambelli, Y.~Demiris, {Multimodal Imitation Using Self-Learned Sensorimotor
  Representations}, in: IEEE/RSJ International Conference on Intelligent Robots
  and Systems, 2016 (2016).

\bibitem{zambelli2016tcds}
M.~Zambelli, Y.~Demiris, {Online Multimodal Ensemble Learning using
  Self-learned Sensorimotor Representations} (2016 (accepted)).

\bibitem{suzuki2016joint}
M.~Suzuki, K.~Nakayama, Y.~Matsuo, Joint multimodal learning with deep
  generative models, arXiv preprint arXiv:1611.01891 (2016).

\bibitem{wu2018multimodal}
M.~Wu, N.~Goodman, Multimodal generative models for scalable weakly-supervised
  learning, in: Advances in Neural Information Processing Systems, 2018, pp.
  5580--5590 (2018).

\bibitem{Droniou2015}
A.~Droniou, S.~Ivaldi, O.~Sigaud, {Deep unsupervised network for multimodal
  perception, representation and classification}, Robotics and Autonomous
  Systems 71 (2015) 83--98 (2015).

\bibitem{higgins2016beta}
I.~Higgins, L.~Matthey, A.~Pal, C.~Burgess, X.~Glorot, M.~Botvinick,
  S.~Mohamed, A.~Lerchner, beta-vae: Learning basic visual concepts with a
  constrained variational framework, in: International Conference on Learning
  Representations, 2016 (2016).

\bibitem{Pickering2014}
M.~J. Pickering, A.~Clark, {Getting ahead: Forward models and their place in
  cognitive architecture}, in: Trends in Cognitive Sciences, Vol.~18, 2014, pp.
  451--456 (2014).

\bibitem{chang2017learning}
H.~J. Chang, T.~Fischer, M.~Petit, M.~Zambelli, Y.~Demiris, Learning kinematic
  structure correspondences using multi-order similarities, IEEE Transactions
  on Pattern Analysis and Machine Intelligence (2017).

\bibitem{tensorflow2015-whitepaper}
M.~Abadi, A.~Agarwal, P.~Barham, E.~Brevdo, Z.~Chen, C.~Citro, G.~S. Corrado,
  A.~Davis, J.~Dean, M.~Devin, S.~Ghemawat, I.~Goodfellow, A.~Harp, G.~Irving,
  M.~Isard, Y.~Jia, R.~Jozefowicz, L.~Kaiser, M.~Kudlur, J.~Levenberg,
  D.~Man\'{e}, R.~Monga, S.~Moore, D.~Murray, C.~Olah, M.~Schuster, J.~Shlens,
  B.~Steiner, I.~Sutskever, K.~Talwar, P.~Tucker, V.~Vanhoucke, V.~Vasudevan,
  F.~Vi\'{e}gas, O.~Vinyals, P.~Warden, M.~Wattenberg, M.~Wicke, Y.~Yu,
  X.~Zheng, \href{http://tensorflow.org/}{{TensorFlow}: Large-scale machine
  learning on heterogeneous systems}, software available from tensorflow.org
  (2015).
\newline\urlprefix\url{http://tensorflow.org/}

\bibitem{maestre2015bootstrapping}
C.~Maestre, A.~Cully, C.~Gonzales, S.~Doncieux, Bootstrapping interactions with
  objects from raw sensorimotor data: a novelty search based approach, in:
  Joint IEEE International Conference on Development and Learning and
  Epigenetic Robotics, IEEE, 2015, pp. 7--12 (2015).

\bibitem{baranes2010intrinsically}
A.~Baranes, P.-Y. Oudeyer, Intrinsically motivated goal exploration for active
  motor learning in robots: A case study, in: IEEE/RSJ International Conference
  on Intelligent Robots and Systems, IEEE, 2010, pp. 1766--1773 (2010).

\end{thebibliography}

\end{document}